\documentclass[11pt]{article}

\usepackage[preprint]{acl}

\usepackage{times}
\usepackage{latexsym}

\usepackage[T1]{fontenc}

\usepackage[utf8]{inputenc}

\usepackage{microtype}

\usepackage{inconsolata}

\usepackage{graphicx}

\usepackage{amsmath}

\usepackage{multirow}

\usepackage{booktabs}
\usepackage{tcolorbox}
\tcbuselibrary{breakable, skins}

\newtcolorbox[auto counter]{promptbox}[2][]{
  colback=gray!5,
  colbacktitle=black!75,
  coltitle=white,
  colframe=black!75,
  boxrule=0.5pt,
  arc=2pt,
  fonttitle=\bfseries\small,
  title={Prompt~\thetcbcounter:~#2},
  #1,
  before upper=\small,
}

\title{On the Tip of the Tongue: First-Token Cueing\\as a Behavioral Probe of Recall Failures in LLMs}

\author{%
Jaewon Sok$^{1}$ \quad
Jewon Yeom$^{2}$ \quad
Seonghyeon Park$^{3}$ \quad
Jeongjae Park$^{2}$ \quad
Hwiyeong Lee$^{2}$ \quad
Taesup Kim$^{2,}$\thanks{\ \ Corresponding author.} \\
$^{1}$Department of Rural Systems Engineering, Seoul National University\\
$^{2}$Graduate School of Data Science, Seoul National University\\
$^{3}$Department of Aerospace Engineering, Seoul National University
}

\begin{document}
\maketitle

\begin{abstract}
Large language models often answer factual questions incorrectly in closed-book
settings, and such errors are usually attributed to missing knowledge. Yet models
often encode answers they fail to produce. We show that many closed-book errors
are first-token recall failures: the model commits to a wrong first token, but
once given the right one it completes the answer and often recognizes it as correct.
We study this with First-Token Cueing, a behavioral probe that prefills the
response with the first token of the gold answer. Model uncertainty peaks at the first answer
token, where it already predicts correctness, and among gold prefixes of any
length the first token alone accounts for the largest share of recovered errors.
Across 16 open models, this one token corrects between a fifth and a half of the
errors on multi-token answers in English QA and improves accuracy in all seven
languages of a multilingual benchmark. Most of the errors it corrects are also
solved by the model's own
sampled reasoning, a criterion for encoded knowledge, and models recognize
correct starts they do not produce: they
continue a correct start more confidently than an incorrect start of the same
answer type and prior probability, yet rarely find it without the gold answer.
These results locate a recoverable class of factual errors at the first decoding
step, where inference-time methods may gain the most.
\end{abstract}

\begin{figure}[t]
    \centering
    \includegraphics[width=\columnwidth]{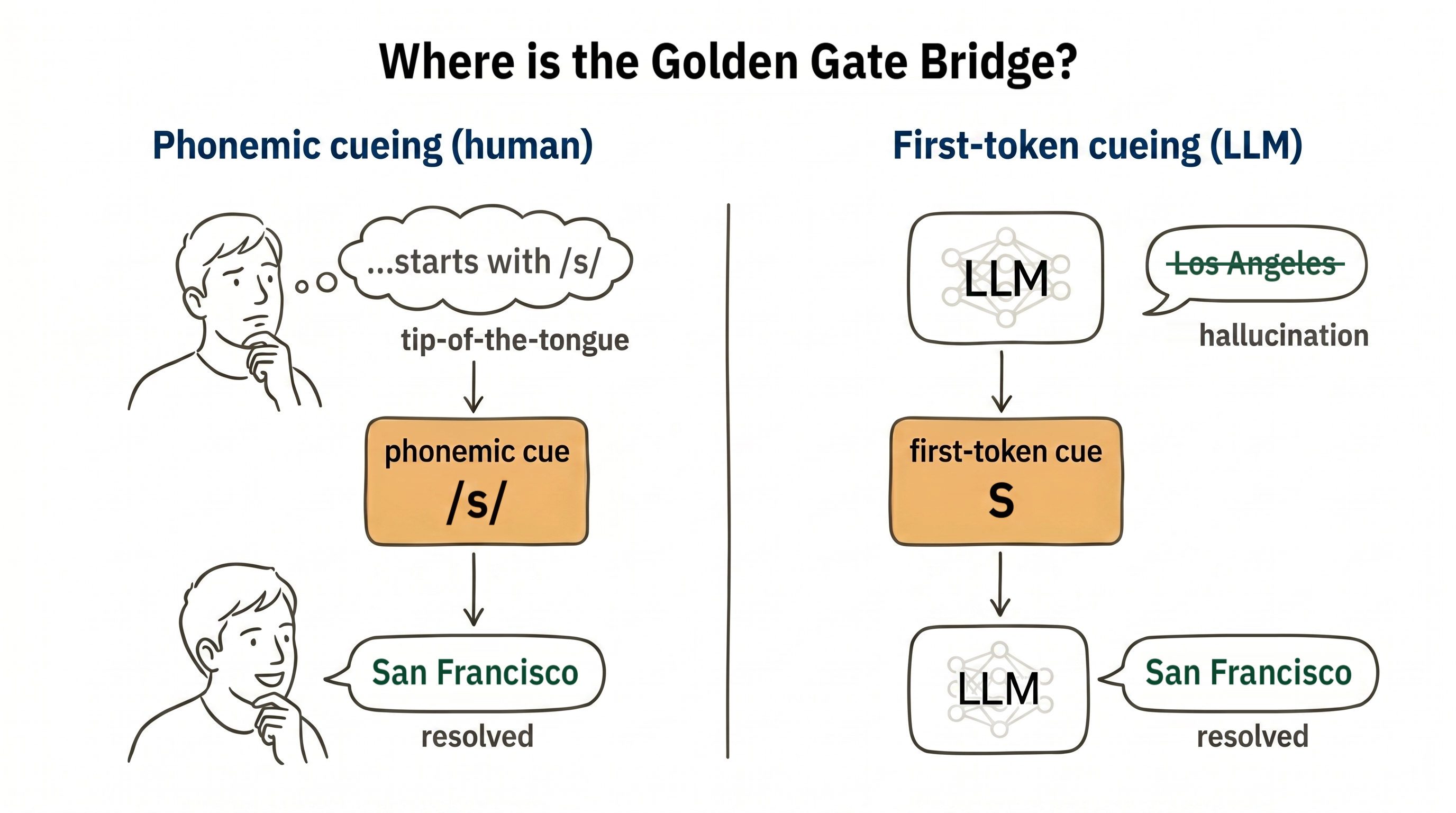}
    \caption{Phonemic cueing in human tip-of-the-tongue states (left;
        \citealp{brown1966tip, james2000phonological}) and First-Token Cueing
        in LLMs (right). In both, production fails at the start of the answer,
        and supplying its first unit, a phoneme or a token, lets the speaker
        or the model complete it.}
    \label{fig:concept}
\end{figure}

\section{Introduction}

Large language models (LLMs) acquire substantial factual knowledge during
pretraining \citep{petroni2019language, roberts2020much}, yet they still make
factual errors in closed-book question answering \citep{ji2023survey,
huang2025survey}. The usual remedy, scaling parameters and data \citep{kaplan2020scaling,
hoffmann2022training}, treats these errors as missing knowledge. Recent work
suggests that many of them are failures of recall instead: probes recover correct
answers from internal states when the generated answer is wrong
\citep{burns2023discovering, azaria2023internal, orgad2025llms}, lightweight
interventions can surface such knowledge \citep{li2023inference,
simhi2024distinguishing}, and models encode many facts they do not verbalize
\citep{gekhman2025inside, calderon2026empty}. Where in generation these failures
arise is less clear.

The first answer token is a natural place to look. In autoregressive
generation, each token is conditioned on the ones before it, so an early
commitment shapes the rest of the answer. The first token of a hallucinated span
carries the strongest error signal \citep{snel2025first}, and answer commitment
localizes to the first decoding step
\citep{yeom2026hallucinationcommitmentfailurelarger}. These studies read the
first token as a signal of error. Whether supplying the correct first token
resolves the error has not been tested.

Human memory offers a parallel. In the tip-of-the-tongue (TOT) state, a speaker
cannot produce a word they know, yet can often report its meaning and parts of
its form, such as its first letter \citep{brown1966tip, schwartz2011tip}. The
transmission deficit hypothesis attributes this to a weak connection between an
available meaning and its sound form, not to missing knowledge
\citep{burke1991tip}. TOT states are often resolved by a partial cue, most
effectively the first sound of the word \citep{meyer1992tip, rastle1996priming,
james2000phonological}, in line with encoding specificity: a cue helps to the
extent that it matches the stored target \citep{tulving1973encoding}. In this
account the cue completes access to a word the speaker already has, rather than
supplying the word (Figure~\ref{fig:concept}). We take this paradigm as a
motivating analogy, not as a claim of a shared mechanism, and ask a behavioral
question: how many closed-book errors of LLMs are resolved by supplying the
start of the answer?

First-Token Cueing prefills the model's response with the first token of the
gold answer and lets the model continue. If many errors are failures of recall,
a cue this small should resolve a large share of them. Because the cue comes
from the gold answer, First-Token Cueing measures which errors are recoverable
rather than fixing them at inference time. The cue also carries information
about the answer, so we call errors it corrects cue-recoverable and test whether
they are recall failures in the sense of prior work, failures to produce an
answer the model encodes.

Our central claim is that many closed-book errors are first-token recall
failures: the model commits to a wrong first token, but once given the right one
it completes the answer and often recognizes it as correct. Three findings across
16 open models from three families support this claim. First, the failure is
located at the first token: uncertainty about correctness concentrates there,
and the first token accounts for the largest share of the errors that gold
prefixes recover. Second, the first token recovers many errors: on multi-token
answers it corrects between a fifth and a half of baseline errors in every model
on the English benchmarks and improves accuracy in every language of a
multilingual benchmark, while random, incorrect, and generic prefixes do not
help. Third, models recognize correct starts they do not produce: when the correct
start lies outside the model's own top candidates, models continue it more
confidently than an incorrect start of the same answer type and prior
probability, yet a search over first characters without the gold answer gains
little because the model prefers its own wrong start.

These results locate a recoverable class of factual errors at the first decoding
step and suggest that inference-time methods should concentrate there.
Our contributions are:
\begin{itemize}
    \item We show that uncertainty about correctness concentrates at the first
    answer token, and that the first token accounts for the largest share of what
    gold prefixes recover (Sections~\ref{sec:locus} and~\ref{sec:k-distribution}).
    \item We show that supplying only the gold first token corrects a large share
    of errors across 16 models and seven languages, and describe how this share
    varies with scale, language, and tokenizer (Section~\ref{sec:recovery}).
    \item We introduce a crossed-cue control that separates recognition from the
    information carried by a cue, and show that models can recognize correct
    answer starts that they fail to produce, but rarely find them on their own
    (Section~\ref{sec:recognition}).
\end{itemize}

\section{Related Work}
\label{sec:related}

\subsection{Recall Failures and Hidden Knowledge in LLMs}

LLMs store many facts in their parameters \citep{petroni2019language,
roberts2020much}, but recall is skewed toward popular facts
\citep{kandpal2023large, mallen2023when} and asymmetric across the direction of a
relation \citep{berglund2024reversal}. Several lines of work find that models
hold more than they produce. Probes read the correct answer from internal states
when the output is wrong \citep{burns2023discovering, azaria2023internal,
orgad2025llms}, and inference-time interventions on these states surface it
\citep{li2023inference, simhi2024distinguishing}. \citet{gekhman2025inside}
measure knowledge as how often a model scores a correct answer above an incorrect
one and find correct answers that the model ranks first yet fails to generate
even with extensive sampling; \citet{calderon2026empty} find that recall rather
than encoding limits frontier models. Our probe is behavioral and needs only
prefilling. Instead of scoring complete answers, it supplies only the first
token, so the model must produce the rest of the answer itself, and our
crossed-cue control compares the correct start with incorrect starts of the same
answer type and prior probability.

\subsection{The First Token in Autoregressive Generation}

Early tokens shape the rest of autoregressive generation. The first token of a
hallucinated span carries the strongest uncertainty signal \citep{snel2025first},
answer commitment localizes to the first decoding step and errors often reflect
probability mass spread over surface forms
\citep{yeom2026hallucinationcommitmentfailurelarger}, and first-token entropy
predicts the difficulty of a reasoning step \citep{zeng2025glimprouter}. These
studies use the first token as a signal; we intervene on it. Branching at the
first decoding step, by continuing each of the most probable first tokens
greedily, uncovers reasoning paths that greedy decoding misses
\citep{wang2024chain}; our first-token oracle (Section~\ref{sec:oracle}) applies
this branching to factual answers. A prefilled start also commits the model:
prefilling a few tokens of a compliant response is enough to bypass safety
alignment \citep{qi2025safety}, and output prefilling has been used to steer the
first token in multiple-choice evaluation \citep{cappelletti2025improving}. Such a generic prefix carries no information
about the answer and, in our open-ended setting, does not help
(Section~\ref{sec:results-oqa}).

\subsection{Tip-of-the-Tongue States in Human Memory}

TOT states were described as early as \citet{james1890principles} and have
been studied experimentally since \citet{brown1966tip}; \citet{schwartz2011tip}
review the field. Models of lexical access separate retrieving a word's meaning
from retrieving its form \citep{dell1986spreading, levelt1999theory}, which the
transmission deficit hypothesis uses to explain TOT states \citep{burke1991tip}.
Feeling-of-knowing judgments show that people recognize information they cannot
produce \citep{hart1965memory}, the human counterpart of the recognition we test
in Section~\ref{sec:recognition}. We adopt the onset-cue paradigm
\citep{meyer1992tip, rastle1996priming, james2000phonological} as a behavioral
test for LLMs. Unlike a phoneme, a token can span several characters or a whole
word, so a first-token cue can carry more of the answer than an onset
(Section~\ref{sec:results-mkqa}).

\begin{figure}[t]
    \centering
    \includegraphics[width=\columnwidth]{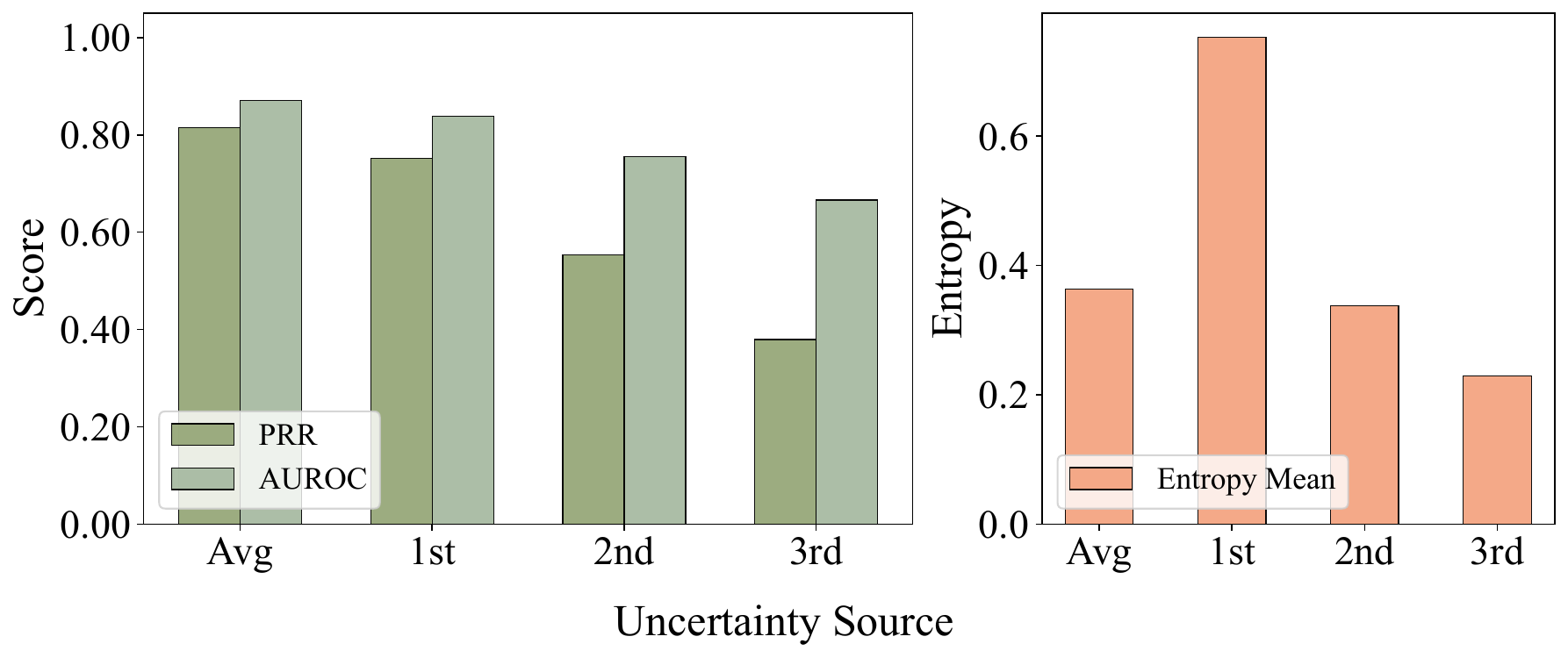}
    \caption{%
        Uncertainty localizes to the first token. AUROC and PRR (left) and mean
        entropy (right) by answer-token position for Llama-3.1 8B on
        TriviaQA. The first token has the highest entropy, and its entropy
        predicts correctness nearly as well as the average over the whole answer.%
    }
    \label{fig:uncertainty}
\end{figure}

\section{Uncertainty Concentrates at the First Token}
\label{sec:locus}

If closed-book errors are failures to start the answer, uncertainty should peak at
the first answer token, and the first token should carry most of the signal about
whether the answer will be correct. Speech production suggests the same:
phonological encoding proceeds left to right \citep{levelt1989speaking,
levelt1999theory}, and in TOT states retrieval stalls at the onset
\citep{brown1966tip, burke1991tip}.

\paragraph{Setup.}
We first examine Llama-3.1 8B on TriviaQA (closed-book, greedy decoding). At each position of the generated answer we
compute the entropy of the next-token distribution, and we compare the first
three positions with the average over all answer tokens. We measure how well each
separates correct from incorrect answers with AUROC and with the Prediction
Rejection Ratio (PRR; \citealp{malinin2021uncertainty}), which compares the
accuracy gained by rejecting the most uncertain answers with that of an oracle
(1: oracle, 0: random).

\paragraph{Uncertainty peaks at the first token.}
Entropy is highest at the first token and drops sharply
afterwards (Figure~\ref{fig:uncertainty}). Lower entropy at later positions is
expected, since the start of an answer constrains its continuation. The
informative result is that first-token entropy alone separates correct from
incorrect answers nearly as well as the whole-answer average, while the second and
third tokens do much worse: the signal about correctness is already present when
the model chooses how to start.

\paragraph{First-token entropy predicts errors in every model.}
For all 16 models, first-token entropy predicts whether the Baseline answer is
correct, with AUROC between about 0.6 and 0.85 on the English benchmarks
(Appendix~\ref{app:rank}). The first-step distribution also predicts which errors
the cue recovers (Section~\ref{sec:recovery}): the cue fixes about two-thirds of
the errors in which the gold first token was the model's second to fifth choice,
and under a third of those in which it ranked below the top thousand. Answer generation thus appears to divide into two
steps:
at the first token the model selects an answer, under high uncertainty that
predicts correctness, and at later tokens it completes the answer under strong
constraint from the prefix. First-Token Cueing intervenes at the first step.

\section{First-Token Cueing}
\label{sec:first-token-cueing}

\paragraph{Formulation.}
Let $q$ be a question with gold answer $y^{\star}$, which the model's tokenizer
$\mathcal{T}$ splits into $\mathbf{t}^{\star} = \mathcal{T}(y^{\star}) =
(t_{1}^{\star}, \ldots, t_{L}^{\star})$. The model's chat template $\Phi$ turns a
system prompt $s$, few-shot demonstrations $\mathcal{F}$, and $q$ into the prompt
$\mathbf{x}_{q} = \Phi(s, \mathcal{F}, q)$, which ends where the model's response
begins.
First-Token Cueing appends the gold first token and lets
$\mathcal{M}_{\theta}$ continue greedily:
\begin{equation}
    \mathbf{c}^{\text{cue}} = \mathbf{x}_{q} \oplus (t_{1}^{\star}), \qquad
    \tilde{\mathbf{y}} = \mathrm{Greedy}\bigl(\mathcal{M}_{\theta}, \mathbf{c}^{\text{cue}}\bigr),
    \label{eq:cue-context}
\end{equation}
where $\oplus$ is concatenation. The evaluated answer is the decoded sequence
$\hat{a} = \mathrm{decode}((t_{1}^{\star}) \oplus \tilde{\mathbf{y}})$.
A $k$-token cue prefills $(t_{1}^{\star}, \ldots, t_{k}^{\star})$ with
$k \leq L$ instead. The Baseline uses the same prompt and decoding without a cue.

\paragraph{Design choices.}
We cue with a token rather than a character or a word because the token is the
unit the model commits to at the first decoding step (Section~\ref{sec:locus}).
Taking the cue from the model's own tokenization of the answer keeps the input in
the form the model expects; a single-character prefill can derail the answer
(Appendix~\ref{app:e2}). The gold answer is the dataset's primary answer
string, and its aliases are used only for grading.

\paragraph{Recall failures and cue-recoverable errors.}
Because $t_{1}^{\star}$ comes from the gold answer, First-Token Cueing is an
oracle intervention that probes errors and cannot be deployed. A recall failure
is an error on a fact the model encodes but does not produce
\citep{gekhman2025inside, calderon2026empty}. Correction by the cue alone does not
show that the fact is encoded, so we call such errors cue-recoverable and test in
Section~\ref{sec:recognition} whether they are recall failures.

\begin{figure*}[!t]
    \centering
    \includegraphics[width=\textwidth]{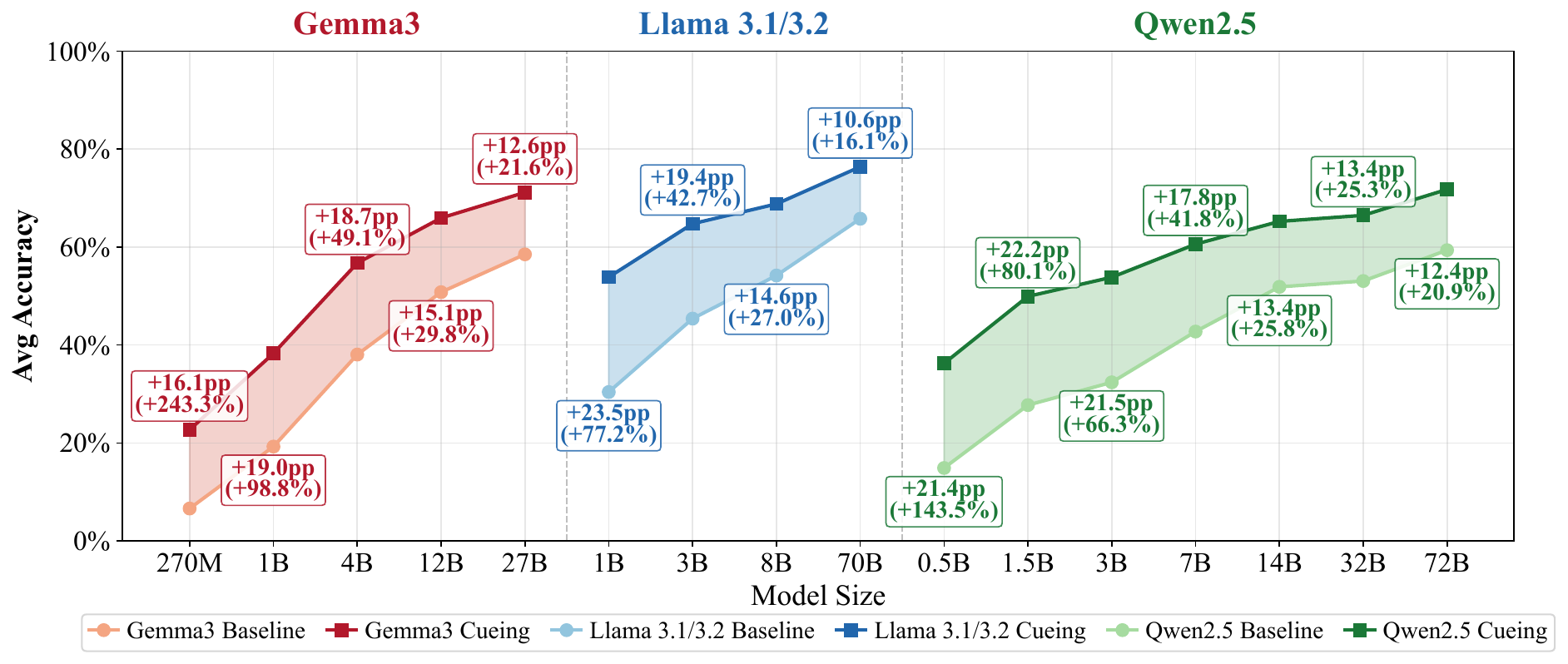}
    \caption{%
        First-Token Cueing improves accuracy for every model. Average accuracy on
        TriviaQA, NQ-Open, and WebQuestions for 16 models, Baseline (light) vs.\
        First-Token Cueing (dark), on questions whose gold answer has at least two
        tokens. Annotations give absolute (pp) and relative (\%) gains. Results on all
        questions are in Figure~\ref{fig:cueing-avg-all}.%
    }
    \label{fig:cueing-avg}
\end{figure*}

\section{The First Token Recovers Many Errors}
\label{sec:recovery}

\subsection{Experimental Setup}
\label{sec:experimental-setup}

\paragraph{Datasets and models.}
We use three English open-domain QA benchmarks, TriviaQA
\citep{joshi2017triviaqa}, NQ-Open \citep{kwiatkowski2019natural}, and
WebQuestions \citep{berant2013semantic}, and the multilingual
MKQA\footnote{\url{https://github.com/apple/ml-mkqa}} \citep{longpre2021mkqa} in
seven languages (English, Spanish, Russian, Arabic, Japanese, Chinese, Korean;
Appendix~\ref{app:datasets}). We evaluate 16 instruction-tuned models from the
Gemma~3, Llama~3.1/3.2, and Qwen~2.5 families, from 270M to 72B parameters,
served with vLLM and decoded greedily
(Appendices~\ref{app:models}--\ref{app:prompts}). All results use the one-token
cue except Section~\ref{sec:k-distribution}, which varies $k$ on the English
benchmarks.

\paragraph{Evaluation.}
A judge model (\texttt{gpt-oss-120b}) grades each response against the gold
answer and its aliases (Appendix~\ref{app:judge}). We report accuracy and the
share of Baseline errors that the cue corrects. A judge might favor a response
that begins with the gold token, but on sampled MKQA responses it agrees with
author-checked labels on 93\% of them and errs mostly by rejecting correct cued
responses (Appendix~\ref{app:judge-validation}). Alias-aware exact match and
token-level F1 \citep{rajpurkar2016squad} also reproduce the gains for every
model and language (Appendix~\ref{app:lexical}).

\paragraph{Headline subset.}
When the gold answer is a single token, the cue is the whole answer. Our headline
results therefore use questions whose gold answer has at least two tokens
($L \geq 2$), which excludes about a tenth of the English questions and fewer of
the MKQA questions. Results on all questions are in
Appendix~\ref{app:all-questions}.

\begin{figure*}[!t]
    \centering
    \includegraphics[width=\textwidth]{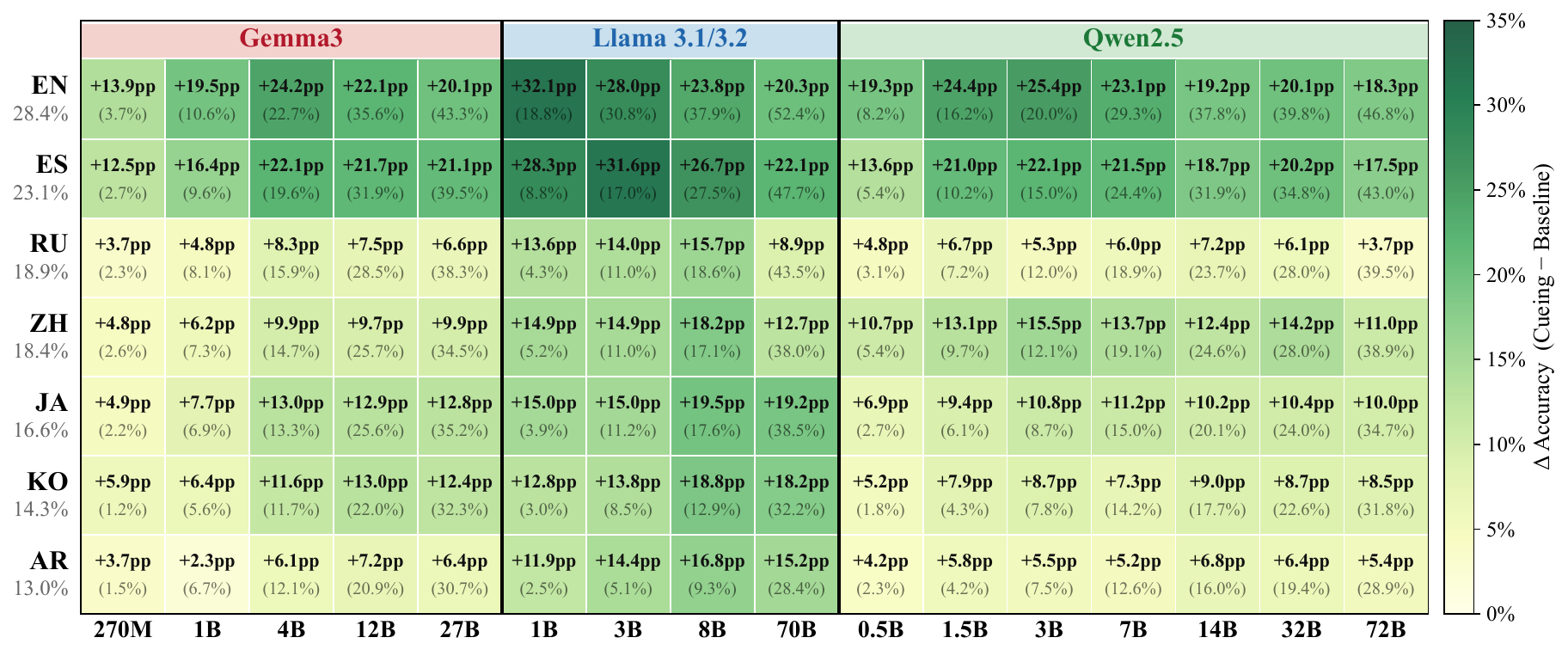}
    \caption{%
        First-Token Cueing improves accuracy in every language for every model on
        MKQA. Each cell shows the gain (Cueing $-$ Baseline, pp) and the Baseline
        accuracy in parentheses, on questions with at least two gold tokens.
        Languages are sorted by mean Baseline accuracy. Results on all questions
        are in Figure~\ref{fig:mkqa-heatmap-all}.%
    }
    \label{fig:mkqa-heatmap}
\end{figure*}

\subsection{English QA}
\label{sec:results-oqa}

\paragraph{The cue corrects many errors.}
Every model improves under First-Token Cueing on every benchmark
(Figure~\ref{fig:cueing-avg}; per-dataset results in
Appendix~\ref{app:full-cueing-results}). On $L \geq 2$ questions, where the model
must produce the rest of the answer itself, the cue corrects between a fifth and
a half of Baseline errors, depending on the model (Table~\ref{tab:fixbreak}).

\paragraph{The gain depends on the identity of the cue.}
Replacing the gold token with a random token lowers accuracy in every model. In
checks on Llama-3.1, a plausible but incorrect first token also lowers accuracy,
and a generic answer prefix \citep{cappelletti2025improving} leaves it unchanged
(Appendices~\ref{app:random-control} and~\ref{app:controls}). Prefilling alone
does not produce the gain. Section~\ref{sec:recognition} examines whether the gain
reflects the model's own knowledge rather than only the information in the cue.

\paragraph{Scale reduces errors but not the share the cue corrects.}
Absolute gains are larger for smaller models: the smallest model of each family
gains roughly 15 to 25 points, and the largest 10 to 13 points. Smaller models,
however, also make more errors. As a share of Baseline errors, the cue corrects
between 40 and 50\% in every model, including the 70B and 72B models, except the
three with the lowest Baseline accuracy (Gemma-3 270M and 1B, Qwen-2.5 0.5B;
Table~\ref{tab:fixbreak}).
Scale reduces the number of errors but not the share the cue recovers, in line
with reports that recall rather than encoding limits
parametric factuality \citep{calderon2026empty}.

\paragraph{The gain holds at every answer length.}
\label{sec:length-analysis}
Including all questions, cueing improves accuracy at every gold answer length in
all 16 models (Appendix~\ref{app:gold-length-buckets}). The gain shrinks as
answers grow longer, consistent with one token specifying less of a longer
answer, but stays positive for answers of five or more tokens.

\subsection{Multilingual QA}
\label{sec:results-mkqa}

\begin{figure}[!t]
    \centering
    \includegraphics[width=\columnwidth]{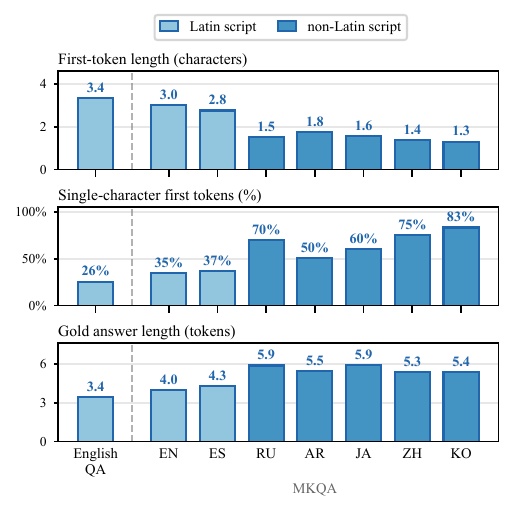}
    \caption{%
        The first token covers less of the answer in non-Latin scripts. Mean
        length of the gold first token in characters (top), share of questions
        whose first token is a single character (middle), and mean gold answer
        length in tokens (bottom), averaged over the Gemma-3, Llama-3, and
        Qwen-2.5 tokenizers. English QA pools TriviaQA, NQ-Open, and
        WebQuestions; per tokenizer in Table~\ref{tab:tokenizer}.%
    }
    \label{fig:tokenizer}
\end{figure}

\paragraph{Cueing helps in every language.}
On MKQA, cueing improves accuracy for every model in every language
(Figure~\ref{fig:mkqa-heatmap}), so cue-recoverable errors are not specific
to English. Gains are largest in English and Spanish and smallest in Arabic and Russian, and
they roughly track Baseline accuracy, with Russian as the clearest exception.

\paragraph{The cue is weaker in non-Latin scripts.}
How much a first token carries depends on the tokenizer and the script
(Figure~\ref{fig:tokenizer}). On the English benchmarks, first tokens span a few
characters under all three tokenizers, and the gain grows with the length of the
first token under each of them while remaining positive when it is a single
character (Appendix~\ref{app:tokenizer}). In the non-Latin scripts, most first
tokens are single characters and answers span more tokens, so a one-token cue
covers less of the answer. Russian illustrates this: its Baseline is
among the highest of the non-Latin languages, yet its gain is among the smallest,
likely because its first token is often a single Cyrillic letter, whereas a single
Chinese or Korean character specifies much more of the answer. This is consistent
with encoding specificity \citep{tulving1973encoding}: a cue helps to the extent
that it specifies the target.

\begin{figure}[!t]
    \centering
    \includegraphics[width=\columnwidth]{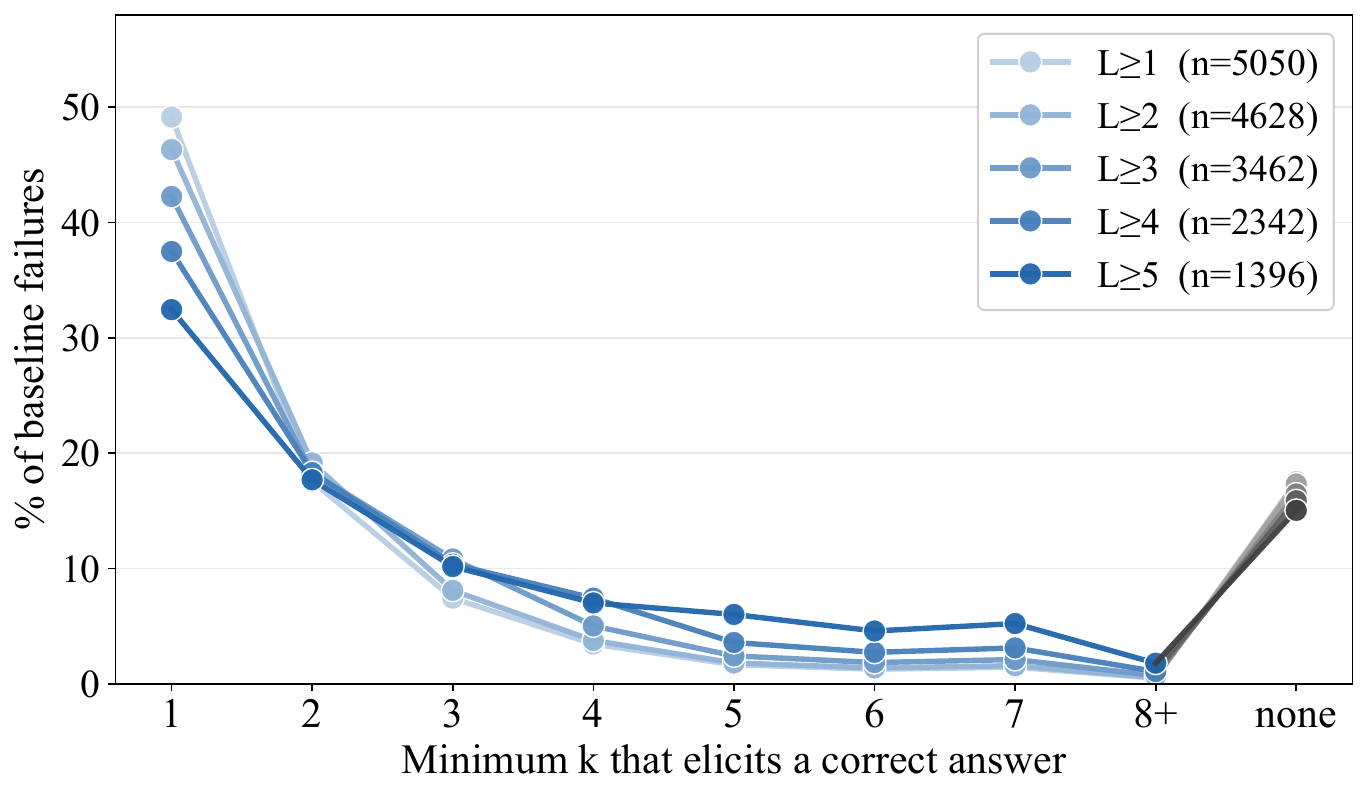}
    \caption{%
        The first token alone recovers more errors than all longer prefixes
        combined. Shortest gold prefix $k$ that makes a Baseline error correct,
        for Llama-3.1 8B. Curve $L \geq \ell$ restricts to gold answers of at least
        $\ell$ tokens; ``none'' marks errors not recovered by the full gold answer.%
    }
    \label{fig:k-distribution}
\end{figure}

\subsection{The First Token Accounts for the Largest Share of Prefix Recovery}
\label{sec:k-distribution}

If the failure lies at the first token, longer gold prefixes should add little
beyond it. For each Baseline error, we find the shortest gold prefix $k$ that
makes the answer correct. For Llama-3.1 8B, the first token alone recovers more errors than
all longer prefixes combined, and $k = 1$ remains the largest group when we
restrict to long answers (Figure~\ref{fig:k-distribution}). The first token is the
largest group at every answer length in all but the three models with the lowest
Baseline accuracy, which need longer prefixes for long answers
(Appendix~\ref{app:k-distribution}).

\section{Models Recognize Correct Starts They Do Not Produce}
\label{sec:recognition}

The gold first token corrects many errors, but it also carries information about
the answer: told that the host city of the 1900 Olympics starts with ``P'', one
can guess ``Paris'' without knowing the fact. We first relate the recovered
errors to what the model produces without a cue (Section~\ref{sec:oracle}). Many
are errors in which the model proposed the correct start but committed to
another, or which its own sampled reasoning also solves, so the cue often selects
an answer the model can reach on its own. We then test the information account
directly (Section~\ref{sec:crossed}). If the cue acted only as information, the
model would continue an incorrect start of the same answer type and prior
probability as confidently as the correct one. It does not, although it rarely finds the correct
start without the gold answer. Unless stated otherwise, we use questions whose gold
answer has at least two tokens.

\begin{table}[!t]
\centering
\resizebox{\columnwidth}{!}{%
\begin{tabular}{llccc}
\toprule
 & & & \multicolumn{2}{c}{Errors recovered (\%)} \\
\cmidrule(lr){4-5}
Model & Stratum & Errors & Cue & Oracle \\
\midrule\midrule
Llama-3.1 8B  & IN     & 2{,}004 (43\%) & 54.0 & 63.0 \\
              & NOT-IN & 2{,}624 (57\%) & 40.4 & 24.8 \\
\midrule
Llama-3.1 70B & IN     & 1{,}265 (41\%) & 49.6 & 57.9 \\
              & NOT-IN & 1{,}793 (59\%) & 44.6 & 22.6 \\
\bottomrule
\end{tabular}%
}
\caption{The cue recovers errors beyond the model's own first-token candidates.
Baseline errors recovered by First-Token Cueing and by the oracle over these
candidates (judge; three English benchmarks pooled; $L \geq 2$). IN: a top-$p$ candidate
is a prefix of the gold first token; NOT-IN: none is. Strict EM in
Table~\ref{tab:oracle-em}.}
\label{tab:oracle}
\end{table}

\subsection{Oracles over the Model's Own Outputs}
\label{sec:oracle}
\label{sec:thinking-comparison}

\paragraph{First-token candidates.}
For each question, we prefill every first token
within the top-$p$ ($p = 0.95$) mass at the first answer position and complete it
greedily. A question is oracle-correct if any of these completions is correct.
The oracle selects with the gold answer, so it upper-bounds any method that
selects among the model's own candidates, such as reranking. We split questions
by whether some candidate, ignoring leading spaces, is a prefix of the gold first
token and can thus be continued into the gold answer (IN) or not (NOT-IN). About two-fifths of the Baseline errors of Llama-3.1 8B and 70B are IN:
the model proposed the correct start and committed to another
(Table~\ref{tab:oracle}). On these errors the cue chooses among the model's own
candidates and recovers about half of them. On NOT-IN errors, cueing recovers
about two-fifths, against about a quarter for the oracle, so the cue reaches
answers that no choice among the model's candidates reaches. This excess remains
when the model has a single candidate and is not largest for first tokens unique
to one answer (Appendices~\ref{app:setsize} and~\ref{app:ambiguity}), so it is
not driven by cues that give the answer away. Section~\ref{sec:crossed} tests
these NOT-IN errors, where the cue is most likely to act only as information.

\begin{figure}[!t]
    \centering
    \includegraphics[width=\columnwidth]{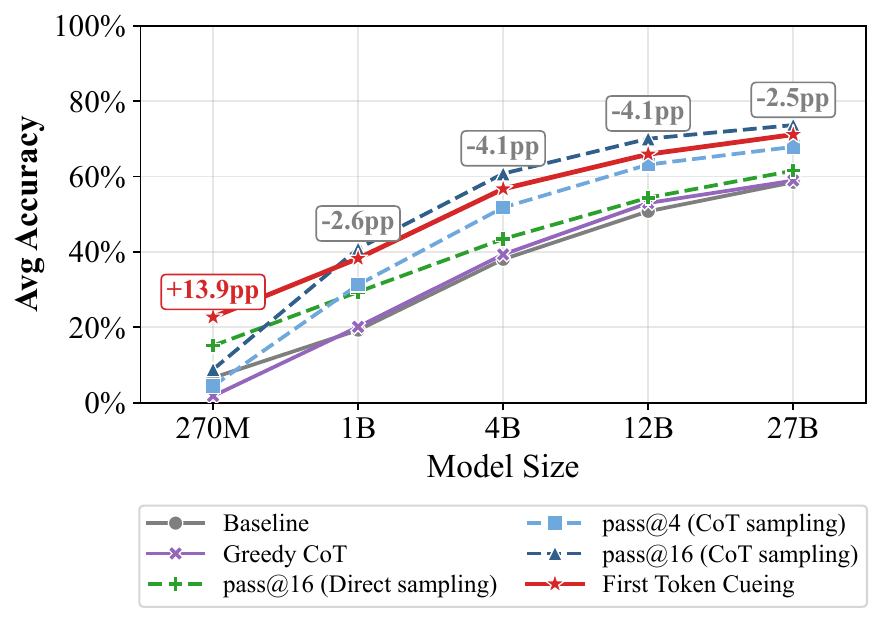}
    \caption{%
        Only sampled chain-of-thought, scored by pass@16, reaches the accuracy
        of First-Token Cueing. Accuracy of First-Token Cueing and five baselines (direct or CoT
        prompt, greedy or sampled decoding) for the Gemma-3 family ($L \geq 2$),
        English QA average. Annotations: Cueing $-$ pass@16 (CoT sampling), pp.%
    }
    \label{fig:thinking-comparison}
\end{figure}

\paragraph{Sampled reasoning.}
Sampled reasoning, which uses no cue, reaches largely the same errors. Pass@16
over chain-of-thought samples \citep{chen2021evaluating}, the criterion by which
\citet{calderon2026empty} mark facts as encoded but not recalled, recovers 77\%
and 62\% of the cue-recoverable errors of Llama-3.1 8B and 70B, against 59\% and
47\% of all their Baseline errors (Appendix~\ref{app:thinking-overlap}). In
accuracy, this pass@16 exceeds cueing by a few points from 1B upward in the
Gemma-3 family, whereas pass@16 over direct answers and greedy chain-of-thought
stay well below cueing (Figure~\ref{fig:thinking-comparison}; Appendix~\ref{app:hparams}). Like the first-token oracle, pass@$k$ selects with
the gold answer, so it shows that the answer is reachable, not that the model can
choose it. The cue shows where reaching it fails: fixing only the first token, it
recovers about three-fifths of the errors that pass@16 marks as encoded (Appendix~\ref{app:thinking-overlap}), so for most of
them the failure lies at the first answer token.

\begin{figure}[!t]
    \centering
    \includegraphics[width=\columnwidth]{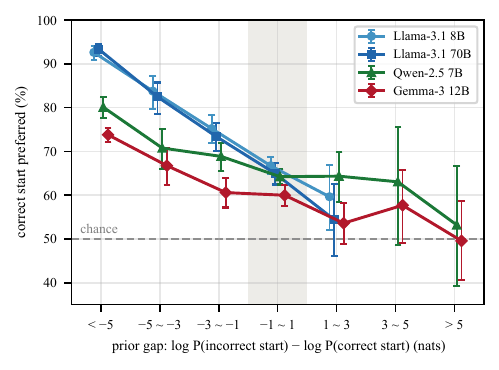}
    \caption{%
        Models continue a correct start more confidently than a type-matched
        incorrect one. Share of NOT-IN questions whose continuation has lower entropy
        after the correct first token than after the incorrect one,
        by the prior gap between the two starts (shaded: matched prior). Error bars: 95\% Wilson intervals.%
    }
    \label{fig:crossed}
\end{figure}

\begin{table}[!t]
\centering
\setlength{\tabcolsep}{4pt}
\resizebox{\columnwidth}{!}{%
\begin{tabular}{lrcccccccc}
\toprule
 & & \multicolumn{2}{c}{Changed} & & & \multicolumn{3}{c}{EM} \\
\cmidrule(lr){3-4}\cmidrule(lr){7-9}
Model & $n$ & Cor. & Inc. & Jacc. & Pref. & Base & Sel. & Cue \\
\midrule\midrule
Llama-3.1 8B  & 2{,}332 & 74 & 78 & 0.76 & 66.7 & 15.3 & 25.3 & 30.1 \\
Llama-3.1 70B & 1{,}485 & 60 & 63 & 0.62 & 65.0 & 26.9 & 29.5 & 35.7 \\
Qwen-2.5 7B   & 2{,}845 & 83 & 89 & 0.82 & 64.2 & 11.6 & 25.3 & 31.1 \\
Gemma-3 12B   & 1{,}760 & 80 & 83 & 0.78 & 60.0 & 14.7 & 25.8 & 34.2 \\
\bottomrule
\end{tabular}%
}
\caption{Models follow either cue but continue the correct one more
confidently. Crossed-cue control at matched prior (the two starts differ by at
most one nat in log-probability; NOT-IN questions; \%). Changed: answer differs from the
Baseline under the correct (Cor.) or incorrect (Inc.) cue; Jacc.: overlap of the
two changed sets; Pref.: the correct-cue continuation has lower mean entropy
(chance 50; 95\% Wilson intervals within $\pm 2.5$; binomial $p < 10^{-16}$ for
every model); EM of the Baseline, of the continuation selected by this
confidence (Sel.), and of the correct cue.}
\label{tab:crossed}
\end{table}

\subsection{Crossed-Cue Control}
\label{sec:crossed}

\paragraph{Design.}
We test whether models respond to a correct start differently from an incorrect
start matched in answer type and prior probability. If the cue only supplies information, a
model asked for an Olympic host city should treat ``P'' (Paris) and ``L''
(London) alike; if it holds the answer, it should treat the correct start
differently. We label each question's answer type with \texttt{gpt-oss-120b}
(seven types such as person, location, and date; Appendix~\ref{app:crossed}) and
pair NOT-IN questions from the same dataset and of the same type, matching the
model's first-token probabilities of the correct and incorrect starts as closely
as possible. Each question receives its own gold first token and its partner's,
which cannot begin a correct answer. Every token is thus the correct cue for one
question and the incorrect cue for the other, which cancels effects of the token
itself, such as tokens whose continuation is fixed regardless of the question
(e.g., \texttt{Leon} $\to$ \texttt{ardo}). We run the control on Llama-3.1 8B and
70B, Qwen-2.5 7B, and Gemma-3 12B, covering nearly all eligible questions, and
report questions whose two starts are matched in prior. For each cue we record
whether the answer changes from the Baseline answer, whether it is correct, and
the confidence of the continuation, its negative mean entropy.

\paragraph{Either cue changes the answer.}
Under either cue, the model changes its Baseline answer on most questions, at
similar rates (Table~\ref{tab:crossed}). The two sets of changed questions
overlap strongly, and nearly all questions changed under both cues are ones the
Baseline answered incorrectly. The incorrect cue almost never yields a correct
answer, and most of its outputs are answers of the right type. Whether the model
abandons its answer thus reflects its uncertainty about the question rather than
the correctness of the start, so the two starts must be compared by how the model
continues them.

\paragraph{Confidence favors the correct start.}
Every model continues the correct start more confidently in about 60 to 67
percent of questions, well above chance (Table~\ref{tab:crossed}). The preference rises with the prior probability of
the correct start and falls to about chance when the incorrect start is far more
probable (Figure~\ref{fig:crossed}). It holds under looser prior matching in all four
models and, for Llama-3.1 8B, under finer type matching and on questions the
Baseline answered incorrectly (Appendix~\ref{app:crossed}). Choosing between the correct and the incorrect start
by this confidence recovers about a third to two-thirds of the gain that the
correct cue alone gives. Models therefore often recognize a
correct start they did not consider, beyond its answer type and prior
probability.

\paragraph{Recognizing a start differs from finding it.}
Without the gold answer, the correct start must also beat the model's own
preferred start. For Llama-3.1 8B, a search over 36 first characters scored by
the model gains about as much as beam search, although two-thirds of questions
have a correct candidate: the correct start is
continued more confidently than most alternatives but usually less so
than the model's own start (Appendix~\ref{app:e2}).

\section{Conclusion}
\label{sec:conclusion}

Using First-Token Cueing, a behavioral probe modeled on onset cueing in
tip-of-the-tongue states, we find that many closed-book errors of 16 open models
are first-token recall failures. Uncertainty about correctness concentrates at
the first answer token. Supplying
only that token corrects a large share of errors on multi-token answers, in every
model on every benchmark and language we tested, and accounts for the largest share of what
longer gold prefixes recover. The errors it corrects are mostly ones the model's
own sampled reasoning also solves, and the recoveries go beyond the information
in the cue: when the correct start lies outside the model's candidates, the model
continues it more confidently than an incorrect start of the same answer type and
prior probability. These errors are failures to start an answer that
the model can complete and recognize once started.

The results suggest where inference-time methods should spend computation:
these errors are decided at the first answer token, whereas decoding and
test-time scaling treat every step alike. Because confidence favors the correct
start over an incorrect one of similar prior probability but not over the model's
own preferred start (Section~\ref{sec:crossed}), choosing among starts will need
evidence from outside the model, such as retrieval, which can also supply starts
outside the model's candidates. Whether
such methods recover the gains of First-Token Cueing without the gold answer
remains open.

\newpage

\section*{Limitations}

We study short-answer factual QA, where the gold answer and its first token are
well defined, which makes the first token a precise target for a behavioral
probe. Extending the probe to long-form generation, where the counterpart would
be the first token of each factual claim, is a natural next step.

First-Token Cueing is a diagnostic rather than a decoding method: it uses the
gold answer to measure how many errors the first token resolves. Accordingly,
we identify recall failures behaviorally, through cue recovery and its overlap
with reasoning-based recovery, rather than from internal states. Supplying the first token
without the gold answer, for example from retrieval, is a direction for future
work.

\section*{Ethical Considerations}
\paragraph{Broader impact.}
Our findings point to the first answer token as a target for inference-time
methods in factual QA, for example by spending more search, sampling, or
retrieval on it when its uncertainty is high. A first token supplied from an
external source is only as reliable as that source, so systems built this way
inherit its errors.

\paragraph{Data and models.}
We use public benchmarks under their licenses (Appendix~\ref{app:datasets}) and
publicly released open-weight models, and we collect no personal data.

\paragraph{Use of AI Assistants.}
In accordance with the ACL Policy on AI Assistance, we
acknowledge the use of Claude Opus 5.5 to assist with
code debugging and improving the readability and grammar
of this manuscript. The final content and responsibility
for the work rest solely with the authors.


\bibliography{custom}

@inproceedings{petroni2019language,
  title     = {Language Models as Knowledge Bases?},
  author    = {Petroni, Fabio and Rockt{\"a}schel, Tim and Riedel, Sebastian
               and Lewis, Patrick and Bakhtin, Anton and Wu, Yuxiang
               and Miller, Alexander H.},
  booktitle = {Proceedings of the 2019 Conference on Empirical Methods in
               Natural Language Processing and the 9th International Joint
               Conference on Natural Language Processing (EMNLP-IJCNLP)},
  pages     = {2463--2473},
  year      = {2019}
}

@inproceedings{roberts2020much,
  title     = {How Much Knowledge Can You Pack Into the Parameters of a
               Language Model?},
  author    = {Roberts, Adam and Raffel, Colin and Shazeer, Noam},
  booktitle = {Proceedings of the 2020 Conference on Empirical Methods in
               Natural Language Processing (EMNLP)},
  pages     = {5418--5426},
  year      = {2020}
}

@article{ji2023survey,
  title={Survey of Hallucination in Natural Language Generation},
   volume={55},
   ISSN={1557-7341},
   url={http://dx.doi.org/10.1145/3571730},
   DOI={10.1145/3571730},
   number={12},
   journal={ACM Computing Surveys},
   publisher={Association for Computing Machinery (ACM)},
   author={Ji, Ziwei and Lee, Nayeon and Frieske, Rita and Yu, Tiezheng and Su, Dan and Xu, Yan and Ishii, Etsuko and Bang, Ye Jin and Madotto, Andrea and Fung, Pascale},
   year={2023},
   month=Mar, pages={1–38}
}

@article{huang2025survey,
  title   = {A Survey on Hallucination in Large Language Models: Principles,
             Taxonomy, Challenges, and Open Questions},
  author  = {Huang, Lei and Yu, Weijiang and Ma, Weitao and Zhong, Weihong
             and Feng, Zhangyin and Wang, Haotian and Chen, Qianglong
             and Peng, Weihua and Feng, Xiaocheng and Qin, Bing and Liu, Ting},
  journal = {ACM Transactions on Information Systems},
  volume  = {43},
  number  = {2},
  pages   = {1--55},
  year    = {2025}
}

@article{kaplan2020scaling,
  title   = {Scaling Laws for Neural Language Models},
  author  = {Kaplan, Jared and McCandlish, Sam and Henighan, Tom
             and Brown, Tom B. and Chess, Benjamin and Child, Rewon
             and Gray, Scott and Radford, Alec and Wu, Jeffrey
             and Amodei, Dario},
  journal = {arXiv preprint arXiv:2001.08361},
  year    = {2020}
}

@inproceedings{hoffmann2022training,
  title     = {Training Compute-Optimal Large Language Models},
  author    = {Hoffmann, Jordan and Borgeaud, Sebastian and Mensch, Arthur
               and Buchatskaya, Elena and Cai, Trevor and Rutherford, Eliza
               and de Las Casas, Diego and Hendricks, Lisa Anne
               and Welbl, Johannes and Clark, Aidan and Hennigan, Tom
               and Noland, Eric and Millican, Katie and van den Driessche, George
               and Damoc, Bogdan and Guy, Aurelia and Osindero, Simon
               and Simonyan, Karen and Elsen, Erich and Rae, Jack W.
               and Vinyals, Oriol and Sifre, Laurent},
  booktitle = {Advances in Neural Information Processing Systems (NeurIPS)},
  year      = {2022}
}

@inproceedings{burns2023discovering,
  title     = {Discovering Latent Knowledge in Language Models Without
               Supervision},
  author    = {Burns, Collin and Ye, Haotian and Klein, Dan and Steinhardt, Jacob},
  booktitle = {The Eleventh International Conference on Learning
               Representations (ICLR)},
  year      = {2023}
}

@inproceedings{azaria2023internal,
  title     = {The Internal State of an {LLM} Knows When It's Lying},
  author    = {Azaria, Amos and Mitchell, Tom},
  booktitle = {Findings of the Association for Computational Linguistics:
               EMNLP 2023},
  pages     = {967--976},
  year      = {2023}
}

@inproceedings{orgad2025llms,
  title     = {{LLM}s Know More Than They Show: On the Intrinsic Representation
               of {LLM} Hallucinations},
  author    = {Orgad, Hadas and Toker, Michael and Gekhman, Zorik
               and Reichart, Roi and Szpektor, Idan and Kotek, Hadas
               and Belinkov, Yonatan},
  booktitle = {The Thirteenth International Conference on Learning
               Representations (ICLR)},
  year      = {2025}
}

@inproceedings{li2023inference,
  title     = {Inference-Time Intervention: Eliciting Truthful Answers from a
               Language Model},
  author    = {Li, Kenneth and Patel, Oam and Vi{\'e}gas, Fernanda
               and Pfister, Hanspeter and Wattenberg, Martin},
  booktitle = {Advances in Neural Information Processing Systems (NeurIPS)},
  year      = {2023}
}

@article{simhi2024distinguishing,
  title   = {Distinguishing Ignorance from Error in {LLM} Hallucinations},
  author  = {Simhi, Adi and Herzig, Jonathan and Szpektor, Idan
             and Belinkov, Yonatan},
  journal = {arXiv preprint arXiv:2410.22071},
  year    = {2024}
}

@inproceedings{gekhman2025inside,
  title     = {Inside-Out: Hidden Factual Knowledge in {LLM}s},
  author    = {Gekhman, Zorik and Ben David, Eyal and Orgad, Hadas
               and Ofek, Eran and Belinkov, Yonatan and Szpektor, Idan
               and Herzig, Jonathan and Reichart, Roi},
  booktitle = {Second Conference on Language Modeling (COLM)},
  year      = {2025}
}

@article{calderon2026empty,
  title   = {Empty Shelves or Lost Keys? Recall Is the Bottleneck for
             Parametric Factuality},
  author  = {Calderon, Nitay and Ben-David, Eyal and Gekhman, Zorik
             and Ofek, Eran and Yona, Gal},
  journal = {arXiv preprint arXiv:2602.14080},
  year    = {2026}
}

@inproceedings{berglund2024reversal,
  title     = {The Reversal Curse: {LLM}s Trained on ``A is B'' Fail to Learn
               ``B is A''},
  author    = {Berglund, Lukas and Tong, Meg and Kaufmann, Max
               and Balesni, Mikita and Stickland, Asa Cooper
               and Korbak, Tomasz and Evans, Owain},
  booktitle = {The Twelfth International Conference on Learning
               Representations (ICLR)},
  year      = {2024}
}

@inproceedings{kandpal2023large,
  title     = {Large Language Models Struggle to Learn Long-Tail Knowledge},
  author    = {Kandpal, Nikhil and Deng, Haikang and Roberts, Adam
               and Wallace, Eric and Raffel, Colin},
  booktitle = {Proceedings of the 40th International Conference on Machine
               Learning (ICML)},
  pages     = {15696--15707},
  year      = {2023}
}

@inproceedings{mallen2023when,
  title     = {When Not to Trust Language Models: Investigating Effectiveness
               of Parametric and Non-Parametric Memories},
  author    = {Mallen, Alex and Asai, Akari and Zhong, Victor and Das, Rajarshi
               and Khashabi, Daniel and Hajishirzi, Hannaneh},
  booktitle = {Proceedings of the 61st Annual Meeting of the Association for
               Computational Linguistics (Volume 1: Long Papers)},
  pages     = {9802--9822},
  year      = {2023}
}

@book{james1890principles,
  title     = {The Principles of Psychology},
  author    = {James, William},
  publisher = {Henry Holt},
  address   = {New York},
  year      = {1890}
}

@article{hart1965memory,
  title   = {Memory and the feeling-of-knowing experience},
  author  = {Hart, John T.},
  journal = {Journal of Educational Psychology},
  volume  = {56},
  number  = {4},
  pages   = {208--216},
  year    = {1965}
}

@article{brown1966tip,
  title   = {The ``tip of the tongue'' phenomenon},
  author  = {Brown, Roger and McNeill, David},
  journal = {Journal of Verbal Learning and Verbal Behavior},
  volume  = {5},
  number  = {4},
  pages   = {325--337},
  year    = {1966}
}

@article{tulving1973encoding,
  title   = {Encoding specificity and retrieval processes in episodic memory},
  author  = {Tulving, Endel and Thomson, Donald M.},
  journal = {Psychological Review},
  volume  = {80},
  number  = {5},
  pages   = {352--373},
  year    = {1973}
}

@article{dell1986spreading,
  title   = {A spreading-activation theory of retrieval in sentence production},
  author  = {Dell, Gary S.},
  journal = {Psychological Review},
  volume  = {93},
  number  = {3},
  pages   = {283--321},
  year    = {1986}
}

@book{levelt1989speaking,
  title     = {Speaking: From Intention to Articulation},
  author    = {Levelt, Willem J. M.},
  publisher = {MIT Press},
  address   = {Cambridge, MA},
  year      = {1989}
}

@article{burke1991tip,
  title   = {On the tip of the tongue: What causes word finding failures in
             young and older adults?},
  author  = {Burke, Deborah M. and MacKay, Donald G. and Worthley, Joanna S.
             and Wade, Elizabeth},
  journal = {Journal of Memory and Language},
  volume  = {30},
  number  = {5},
  pages   = {542--579},
  year    = {1991}
}

@article{meyer1992tip,
  title   = {The tip-of-the-tongue phenomenon: Blocking or partial activation?},
  author  = {Meyer, Antje S. and Bock, Kathryn},
  journal = {Memory \& Cognition},
  volume  = {20},
  number  = {6},
  pages   = {715--726},
  year    = {1992}
}

@article{rastle1996priming,
  title   = {Priming the tip of the tongue: Effects of prior processing on
             word retrieval in young and older adults},
  author  = {Rastle, Kathleen G. and Burke, Deborah M.},
  journal = {Journal of Memory and Language},
  volume  = {35},
  number  = {4},
  pages   = {586--605},
  year    = {1996}
}

@article{levelt1999theory,
  title   = {A Theory of Lexical Access in Speech Production},
  author  = {Levelt, Willem J. M. and Roelofs, Ardi and Meyer, Antje S.},
  journal = {Behavioral and Brain Sciences},
  volume  = {22},
  number  = {1},
  pages   = {1--38},
  year    = {1999}
}

@article{james2000phonological,
  title   = {Phonological priming effects on word retrieval and
             tip-of-the-tongue experiences in young and older adults},
  author  = {James, Lori E. and Burke, Deborah M.},
  journal = {Journal of Experimental Psychology: Learning, Memory, and
             Cognition},
  volume  = {26},
  number  = {6},
  pages   = {1378--1391},
  year    = {2000}
}

@article{schwartz2011tip,
  title   = {Tip-of-the-tongue ({TOT}) states: Retrieval, behavior, and experience},
  author  = {Schwartz, Bennett L. and Metcalfe, Janet},
  journal = {Memory \& Cognition},
  volume  = {39},
  number  = {5},
  pages   = {737--749},
  year    = {2011}
}

@article{snel2025first,
  title   = {First Hallucination Tokens Are Different from Conditional Ones},
  author  = {Snel, Jakob and Oh, Seong Joon},
  journal = {arXiv preprint arXiv:2507.20836},
  year    = {2025}
}

@misc{zeng2025glimprouter,
  title={GlimpRouter: Efficient Collaborative Inference by Glimpsing One Token of Thoughts}, 
  author={Wenhao Zeng and Xuteng Zhang and Yuling Shi and Chao Hu and Yuting Chen and Beijun Shen and Xiaodong Gu},
  year={2026},
  eprint={2601.05110},
  archivePrefix={arXiv},
  primaryClass={cs.AI},
  url={https://arxiv.org/abs/2601.05110}, 
}

@misc{yeom2026hallucinationcommitmentfailurelarger,
      title={Hallucination as Commitment Failure: Larger {LLMs} Misfire Despite Knowing the Answer}, 
      author={Jewon Yeom and Jaewon Sok and Heejun Kim and Seonghyeon Park and Jeongjae Park and Taesup Kim},
      year={2026},
      eprint={2605.22007},
      archivePrefix={arXiv},
      primaryClass={cs.CL},
      url={https://arxiv.org/abs/2605.22007}, 
}

@inproceedings{joshi2017triviaqa,
  title     = {{TriviaQA}: A Large Scale Distantly Supervised Challenge
               Dataset for Reading Comprehension},
  author    = {Joshi, Mandar and Choi, Eunsol and Weld, Daniel S.
               and Zettlemoyer, Luke},
  booktitle = {Proceedings of the 55th Annual Meeting of the Association
               for Computational Linguistics (Volume 1: Long Papers)},
  pages     = {1601--1611},
  year      = {2017}
}

@article{kwiatkowski2019natural,
  title = "Natural Questions: A Benchmark for Question Answering Research",
    author = "Kwiatkowski, Tom  and
      Palomaki, Jennimaria  and
      Redfield, Olivia  and
      Collins, Michael  and
      Parikh, Ankur  and
      Alberti, Chris  and
      Epstein, Danielle  and
      Polosukhin, Illia  and
      Devlin, Jacob  and
      Lee, Kenton  and
      Toutanova, Kristina  and
      Jones, Llion  and
      Kelcey, Matthew  and
      Chang, Ming-Wei  and
      Dai, Andrew M.  and
      Uszkoreit, Jakob  and
      Le, Quoc  and
      Petrov, Slav",
    editor = "Lee, Lillian  and
      Johnson, Mark  and
      Roark, Brian  and
      Nenkova, Ani",
    journal = "Transactions of the Association for Computational Linguistics",
    volume = "7",
    year = "2019",
    address = "Cambridge, MA",
    publisher = "MIT Press",
    url = "https://aclanthology.org/Q19-1026/",
    doi = "10.1162/tacl_a_00276",
    pages = "452--466"

}

@inproceedings{berant2013semantic,
  title     = {Semantic Parsing on {Freebase} from Question-Answer Pairs},
  author    = {Berant, Jonathan and Chou, Andrew and Frostig, Roy
               and Liang, Percy},
  booktitle = {Proceedings of the 2013 Conference on Empirical Methods
               in Natural Language Processing (EMNLP)},
  pages     = {1533--1544},
  year      = {2013}
}

@article{longpre2021mkqa,
  title   = {{MKQA}: A Linguistically Diverse Benchmark for Multilingual
             Open Domain Question Answering},
  author  = {Longpre, Shayne and Lu, Yi and Daiber, Joachim},
  journal = {Transactions of the Association for Computational Linguistics},
  volume  = {9},
  pages   = {1389--1406},
  year    = {2021}
}

@misc{chen2021evaluating,
  title={Evaluating Large Language Models Trained on Code}, 
  author={Mark Chen and Jerry Tworek and Heewoo Jun and Qiming Yuan and Henrique Ponde de Oliveira Pinto and Jared Kaplan and Harri Edwards and Yuri Burda and Nicholas Joseph and Greg Brockman and Alex Ray and Raul Puri and Gretchen Krueger and Michael Petrov and Heidy Khlaaf and Girish Sastry and Pamela Mishkin and Brooke Chan and Scott Gray and Nick Ryder and Mikhail Pavlov and Alethea Power and Lukasz Kaiser and Mohammad Bavarian and Clemens Winter and Philippe Tillet and Felipe Petroski Such and Dave Cummings and Matthias Plappert and Fotios Chantzis and Elizabeth Barnes and Ariel Herbert-Voss and William Hebgen Guss and Alex Nichol and Alex Paino and Nikolas Tezak and Jie Tang and Igor Babuschkin and Suchir Balaji and Shantanu Jain and William Saunders and Christopher Hesse and Andrew N. Carr and Jan Leike and Josh Achiam and Vedant Misra and Evan Morikawa and Alec Radford and Matthew Knight and Miles Brundage and Mira Murati and Katie Mayer and Peter Welinder and Bob McGrew and Dario Amodei and Sam McCandlish and Ilya Sutskever and Wojciech Zaremba},
  year={2021},
  eprint={2107.03374},
  archivePrefix={arXiv},
  primaryClass={cs.LG},
  url={https://arxiv.org/abs/2107.03374}, 
}

@article{cappelletti2025improving,
  title={Improving {LLM} First-Token Predictions in Multiple-Choice Question Answering via Output Prefilling},
  author={Cappelletti, Silvia and Poppi, Tobia and Poppi, Samuele and Yong, Zheng-Xin and Garcia-Olano, Diego and Cornia, Marcella and Baraldi, Lorenzo and Cucchiara, Rita},
  journal={arXiv preprint arXiv:2505.15323},
  year={2025}
}

@inproceedings{rajpurkar2016squad,
  title={{SQuAD}: 100,000+ Questions for Machine Comprehension of Text},
  author={Rajpurkar, Pranav and Zhang, Jian and Lopyrev, Konstantin and Liang, Percy},
  booktitle={Proceedings of the 2016 Conference on Empirical Methods in Natural Language Processing},
  pages={2383--2392},
  year={2016}
}

@inproceedings{wang2024chain,
  title     = {Chain-of-Thought Reasoning Without Prompting},
  author    = {Wang, Xuezhi and Zhou, Denny},
  booktitle = {Advances in Neural Information Processing Systems (NeurIPS)},
  year      = {2024}
}

@inproceedings{qi2025safety,
  title     = {Safety Alignment Should Be Made More Than Just a Few Tokens Deep},
  author    = {Qi, Xiangyu and Panda, Ashwinee and Lyu, Kaifeng and Ma, Xiao
               and Roy, Subhrajit and Beirami, Ahmad and Mittal, Prateek
               and Henderson, Peter},
  booktitle = {The Thirteenth International Conference on Learning Representations (ICLR)},
  year      = {2025}
}

@inproceedings{malinin2021uncertainty,
  title     = {Uncertainty Estimation in Autoregressive Structured Prediction},
  author    = {Malinin, Andrey and Gales, Mark},
  booktitle = {International Conference on Learning Representations (ICLR)},
  year      = {2021}
}

\clearpage
\appendix
\renewcommand{\topfraction}{0.8}
\renewcommand{\dbltopfraction}{0.75}
\renewcommand{\textfraction}{0.2}
\renewcommand{\floatpagefraction}{0.7}
\renewcommand{\dblfloatpagefraction}{0.7}

\section{Experimental Details}
\label{app:experimental-details}

\begin{table*}[!t]
\centering
\small
\begin{tabular}{p{0.13\textwidth} p{0.06\textwidth} p{0.55\textwidth} p{0.12\textwidth}}
\toprule
Dataset & Year & Description & License \\
\midrule\midrule
TriviaQA \citep{joshi2017triviaqa} & 2017 &
Open-domain trivia questions with Wikipedia-grounded short answers. We
use the \texttt{rc.wikipedia.nocontext} subset, validation split. &
Apache 2.0 \\
\midrule
NQ-Open \citep{kwiatkowski2019natural} & 2019 &
Open-domain variant of Natural Questions: real Google search queries
paired with short Wikipedia-grounded answers. Validation split. &
CC BY-SA 3.0 \\
\midrule
WebQuestions \citep{berant2013semantic} & 2013 &
Real-world questions issued to the Google Suggest API, paired with
Freebase-grounded short answers. Test split. &
CC BY 4.0 \\
\midrule
MKQA \citep{longpre2021mkqa} & 2021 &
Multilingual Knowledge Questions and Answers: NQ-Open queries translated
and answered in 26 languages. We use 7 languages (en, es, ru, ar,
ja, zh\_cn, ko). &
CC BY-SA 3.0 \\
\bottomrule
\end{tabular}
\caption{Datasets used in this study, with year of publication, description,
split, and license. All datasets are used in accordance with their stated
license.}
\label{tab:datasets}
\end{table*}

\subsection{Datasets}
\label{app:datasets}

Table~\ref{tab:datasets} summarizes the datasets used in this study.
All datasets are loaded through the HuggingFace \texttt{datasets}
library, except MKQA, which is read from a local JSONL file. For
NQ-Open and WebQuestions, each item provides a list of acceptable
answers; we treat the first entry as the gold answer $y^{\star}$ and
the remainder as aliases $\mathcal{A}(y^{\star})$. For TriviaQA and
MKQA, the provided alias list is used directly. For MKQA, samples
whose answer string for the target language is empty are filtered out.

\subsection{Models and Inference Backend}
\label{app:models}

We evaluate $16$ open-weight instruction-tuned models from three
dense-Transformer families: five Gemma-3 models ($270$M, $1$B, $4$B,
$12$B, and $27$B parameters), seven Qwen-2.5 models ($0.5$B, $1.5$B,
$3$B, $7$B, $14$B, $32$B, and $72$B), and four Llama models spanning
two generations (Llama-3.2 at $1$B and $3$B; Llama-3.1 at $8$B and
$70$B). All checkpoints are the official HuggingFace ``-Instruct''
(Qwen, Llama) or ``-it'' (Gemma) variants.

All models are served with vLLM on a single NVIDIA B200 GPU (no
tensor parallelism), using a context window of $4{,}096$ tokens,
bfloat16 precision, and GPU memory utilization tuned per model around
$0.9$.

For each model, prompts are wrapped in the model's native chat
template with a generation-prompt marker appended, so that all inputs
are tokenized in the model's expected chat format rather than as raw
text.

\subsection{Decoding Hyperparameters}
\label{app:hparams}

We compare six conditions that vary along two axes: prompt
(direct-answer vs.\ chain-of-thought) and decoding (greedy vs.\
sampling). Table~\ref{tab:hparams} summarizes the hyperparameters.

Baseline and First-Token Cueing share the same
decoding policy (greedy, $50$-token budget) and the direct-answer
prompt; they differ only in whether the gold answer's first token
is prepended to the assistant response. Greedy CoT uses
the CoT prompt with greedy decoding and a $1{,}024$-token budget
to accommodate the reasoning trace. The three sampling conditions
all use temperature $0.7$ and top-$p = 0.95$ with $n = 16$
generations per question: pass@16 (Direct sampling) uses
the direct-answer prompt with a $50$-token budget;
pass@$k$ (CoT sampling) ($k \in \{4, 16\}$) uses the CoT
prompt with a $1{,}024$-token budget and shares a single $n = 16$
sample pool, with pass@$k$ computed via the unbiased estimator of
\citet{chen2021evaluating}.

\begin{table}[!t]
\centering
\resizebox{\columnwidth}{!}{%
\begin{tabular}{lccccc}
\toprule
Condition & Prompt & Max tokens & Temp. & Top-$p$ & $n$ \\
\midrule\midrule
Baseline & Direct & 50 & 0.0 & -- & 1 \\
First-Token Cueing & Direct + cue & 50 & 0.0 & -- & 1 \\
Greedy CoT & CoT & 1024 & 0.0 & -- & 1 \\
pass@16 (Direct sampling) & Direct & 50 & 0.7 & 0.95 & 16 \\
pass@$k$ (CoT sampling) & CoT & 1024 & 0.7 & 0.95 & 16 \\
\bottomrule
\end{tabular}%
}
\caption{Decoding hyperparameters per condition. ``Direct'' uses
Prompt~\ref{prompt:direct-system}; ``CoT'' uses
Prompt~\ref{prompt:CoT-system}. ``Direct + cue'' denotes the
direct-answer prompt with the gold answer's first token prepended
to the assistant response. pass@$k$ (CoT sampling) shares the same
$n = 16$ sample pool for $k \in \{4, 16\}$.}
\label{tab:hparams}
\end{table}

\subsection{Prompt Templates}
\label{app:prompts}

\paragraph{Direct-answer prompt.}
Used by Baseline, First-Token Cueing, and pass@16 (Direct sampling).
A fixed system message (Prompt~\ref{prompt:direct-system}) is
followed by five fixed user/assistant demonstrations, with the
current question $q$ appended as the final user turn. Applying
$\Phi(\cdot)$ to this message sequence yields the prefix
$\mathbf{x}_{q}$.

\begin{promptbox}[label=prompt:direct-system]{Direct-answer System Prompt}
Answer the question directly. Provide only the exact answer without
any explanation or conversational filler.
\end{promptbox}

\paragraph{CoT prompt.}
Used by Greedy CoT, pass@4 (CoT sampling), and pass@16 (CoT
sampling). Zero-shot chain-of-thought with no demonstrations.

\begin{promptbox}[label=prompt:CoT-system]{CoT System Prompt}
You are a careful and logical thinker. First, think step-by-step
about the question. After your reasoning, you must conclude your
response on a new line with the exact phrase: ``The answer is
[your exact short answer]''.
\end{promptbox}

The short answer is extracted from the model's full output by
case-insensitive matching of the phrase ``the answer is'', allowing
optional whitespace and separator characters (colon, hyphen, or
asterisk) before the answer span, with the remainder captured as
the parsed answer. On parsing failure, the full output is still
forwarded to the judge model, marked as ``[Parsing Failed]''.

\subsection{LLM-as-Judge}
\label{app:judge}

All responses are graded by a single judge model, \texttt{gpt-oss-120b},
served with vLLM in eager-execution mode with chunked prefill disabled,
a context window of $4{,}096$ tokens, and GPU memory utilization set to
$0.80$. Decoding is greedy. The judge system prompt is given in
Prompt~\ref{prompt:judge-system}, with condition-specific user prompts
in Prompts~\ref{prompt:judge-direct} and~\ref{prompt:judge-CoT}.

\begin{promptbox}[label=prompt:judge-system]{Judge System Prompt}
You are an objective expert evaluator. Your task is to determine if the
Model Output (or the provided Parsed Final Answer) contains or is
functionally equivalent to the ``Correct Answers''. Ignore minor
differences in punctuation, casing, articles, or extra conversational
words. If the Model Output contains the correct information alongside
extra details, consider it correct.

\medskip
\noindent CRITICAL RULE: First, write your reasoning. Then,
conclude your response on a new line with the exact phrase: ``The
answer is True'' or ``The answer is False''.
\end{promptbox}

\begin{promptbox}[label=prompt:judge-direct]{Judge User Prompt (Baseline \& First-Token Cueing)}
Question: \{question\} \\
Correct Answers: \{golden\_answer\}, \{alias\_1\}, \{alias\_2\}, \ldots \\
Model Output: \{model\_output\} \\[4pt]
Is the Model Output correct? Write your reasoning, and finish with
``The answer is True'' or ``The answer is False''.
\end{promptbox}

\begin{promptbox}[label=prompt:judge-CoT]{Judge User Prompt (CoT)}
Question: \{question\} \\
Correct Answers: \{golden\_answer\}, \{alias\_1\}, \{alias\_2\}, \ldots \\[2pt]
[Model Full Output]: \\
\{full\_cot\_output\} \\[2pt]
[Parsed Final Answer]: \{parsed\_answer\} \\[4pt]
Is the Model Output / Parsed Final Answer correct? Write your reasoning,
and finish with ``The answer is True'' or ``The answer is False''.
\end{promptbox}

\paragraph{Verdict parsing.}
The judge's verdict is reduced to a Boolean by the following priority.
First, we look for the case-insensitive phrase ``the answer is''
followed by ``true'' or ``false'', with optional whitespace or colons
in between. If no such match is found, we search for variant keywords
such as ``final true'', ``is true'', ``final false'', or ``is false''.
As a final fallback, we compare the positions of the last occurrence
of ``true'' versus ``false'' in the judge's response.

\subsection{Judge Agreement with Author-Checked Labels}
\label{app:judge-validation}

\begin{table}[t]
\centering
\resizebox{\columnwidth}{!}{%
\begin{tabular}{lcccccc}
\toprule
& \multicolumn{3}{c}{Sample} & \multicolumn{3}{c}{Reweighted to $L \geq 2$} \\
\cmidrule(lr){2-4}\cmidrule(lr){5-7}
& Agree. & $\kappa$ & FP / FN & Agree. & Gain$_{\mathrm{judge}}$ & Gain$_{\mathrm{labels}}$ \\
\midrule\midrule
EN & 90.0 & 0.80 & 0 / 5 & 91.5 & $+23.8$ & $+30.4$ \\
KO & 92.0 & 0.84 & 1 / 3 & 90.6 & $+18.8$ & $+23.0$ \\
JA & 98.0 & 0.96 & 0 / 1 & 96.6 & $+19.5$ & $+12.7$ \\
ZH & 92.0 & 0.84 & 0 / 4 & 89.1 & $+18.2$ & $+26.2$ \\
\midrule
Baseline & 95.0 & 0.90 & 1 / 4 & 93.2 & -- & -- \\
Cue & 91.0 & 0.82 & 0 / 9 & 90.7 & -- & -- \\
\midrule
All & 93.0 & 0.86 & 1 / 13 & 91.9 & $+20.1$ & $+23.1$ \\
\bottomrule
\end{tabular}%
}
\caption{The judge agrees with author-checked labels and errs toward rejecting
correct responses. Llama-3.1 8B on MKQA ($L \geq 2$); 50 responses per language,
half Baseline and half First-Token Cueing, stratified by the judge's verdict.
Agree.: agreement with the labels (\%). FP: the judge accepts an incorrect
response; FN: it rejects a correct one. Reweighted columns weight each verdict
stratum by its share in the full $L \geq 2$ subset; Gain: Cue $-$ Baseline
accuracy (points) under the judge and under the labels, averaged over languages
in the last row.}
\label{tab:judge-validation}
\end{table}

\paragraph{Sample.}
We check the judge on Llama-3.1 8B responses to MKQA in English, Korean,
Japanese, and Chinese. For each language, we draw 50 distinct questions from the
$L \geq 2$ subset, 25 with the Baseline response and 25 with the First-Token
Cueing response. The sample is stratified by the judge's verdict: the judge
accepts 13 of the Baseline and 12 of the cued responses in each language and
rejects the rest.

\paragraph{Labels.}
A response is labeled correct if it is semantically equivalent to the gold
answer or one of its aliases. Translations, transliterations, and script
variants count, as does a correct answer accompanied by extra content, and a
list that includes the gold answer when the question asks for several entities.
An author checked all 200 labels.

\paragraph{Agreement.}
The judge agrees with the labels on 93.0\% of the responses (95\% Wilson
interval 88.6--95.8; Cohen's $\kappa = 0.86$; Table~\ref{tab:judge-validation}).
Of the 14 disagreements, 13 are correct responses that the judge rejects, so the
judge is conservative: 99\% of the responses it accepts are correct, whereas 87\%
of those it rejects are incorrect. It rejects five responses that give only the
year for a date (``1776'' for 1776-07-04), three correct answers followed by
other entities (``United Kingdom and Ireland'' for United Kingdom), two in which
it treats the aliases as jointly required (``Lonnie Donegan'' for Lonnie Donegan
with alias Anthony James Donegan), two surface-form variants (``Fraternals'' for
fraternal twins), and one rounded number (``114,000'' for 114,194). It accepts
one response that does not answer the question. Counting year-only answers as
incorrect raises agreement to 95.5\%.

\paragraph{Cued responses.}
The judge does not favor cued responses. It accepts no incorrect cued response,
and nine of its 13 rejections of correct responses are cued responses, including
all three with extra entities. Agreement is 95.0\% on Baseline and 91.0\% on
cued responses ($\kappa = 0.90$ and $0.82$).

\paragraph{Population estimates.}
Because the sample is stratified by the judge's verdict, we weight each verdict
stratum by its share in the full $L \geq 2$ subset of its language and
condition, and draw 95\% intervals from a Jeffreys posterior for each stratum
($10{,}000$ draws). The reweighted agreement averages 91.9\%. Averaged over
languages, the judge underestimates accuracy under the labels by 6.3 points for
the Baseline and by 9.3 points under the cue, so it understates the gain: $+20.1$
points under the judge against an estimated $+23.1$ points under the labels (95\%
interval $+11.8$ to $+29.9$). Strata hold 12 or 13 responses, so per-language
estimates are noisy. A single flipped label moves a cell by about seven points,
which accounts for Japanese, the only language in which the estimated gain falls
below the judge's.

\section{Complete First-Token Cueing Results}
\label{app:full-cueing-results}

\begin{table}[!htb]
\centering
\resizebox{\columnwidth}{!}{%
\begin{tabular}{lrrrrr}
\toprule
Model & $n$ & Errors & Corrected (\%) & Broken (\%) & Ratio \\
\midrule\midrule
Gemma-3 270M & 11{,}593 & 10{,}875 & 21.6 & 37.2 & 8.8 \\
Gemma-3 1B & 11{,}593 & 9{,}253 & 32.7 & 18.8 & 6.9 \\
Gemma-3 4B & 11{,}593 & 6{,}623 & 41.4 & 9.3 & 5.9 \\
Gemma-3 12B & 11{,}593 & 4{,}829 & 43.3 & 6.4 & 4.9 \\
Gemma-3 27B & 11{,}593 & 3{,}856 & 44.6 & 4.9 & 4.6 \\
\midrule
Llama-3.2 1B & 12{,}253 & 8{,}160 & 44.2 & 12.4 & 7.1 \\
Llama-3.2 3B & 12{,}253 & 5{,}936 & 48.6 & 8.1 & 5.6 \\
Llama-3.1 8B & 12{,}253 & 4{,}628 & 46.3 & 5.5 & 5.1 \\
Llama-3.1 70B & 12{,}253 & 3{,}058 & 46.7 & 4.2 & 3.7 \\
\midrule
Qwen-2.5 0.5B & 12{,}411 & 10{,}554 & 31.9 & 19.3 & 9.4 \\
Qwen-2.5 1.5B & 12{,}411 & 8{,}653 & 40.5 & 12.2 & 7.6 \\
Qwen-2.5 3B & 12{,}411 & 7{,}916 & 42.4 & 11.6 & 6.4 \\
Qwen-2.5 7B & 12{,}411 & 6{,}519 & 43.5 & 7.7 & 6.2 \\
Qwen-2.5 14B & 12{,}411 & 5{,}103 & 43.0 & 7.0 & 4.3 \\
Qwen-2.5 32B & 12{,}411 & 4{,}913 & 42.9 & 6.2 & 4.5 \\
Qwen-2.5 72B & 12{,}411 & 4{,}006 & 44.5 & 4.8 & 4.4 \\
\bottomrule
\end{tabular}%
}
\caption{Errors corrected and answers broken by First-Token Cueing on English QA
($L \geq 2$; TriviaQA, NQ-Open, and WebQuestions pooled; judge). $n$: questions;
Errors: Baseline errors; Corrected: share of Baseline errors that the cue makes
correct; Broken: share of Baseline-correct answers that the cue makes incorrect;
Ratio: corrected over broken answers.}
\label{tab:fixbreak}
\end{table}

First-Token Cueing improves over Baseline in all $160$ (model, task) cells of the
headline subset, questions whose gold answer has at least two tokens under the
model's tokenizer (Table~\ref{tab:L2-cueing-results}). The same holds on all
questions (Appendix~\ref{app:all-questions}).

\begin{table*}[t]
\centering
\footnotesize
\setlength{\tabcolsep}{3pt}
\renewcommand{\arraystretch}{1.0}
\resizebox{\textwidth}{!}{%
\begin{tabular}{llccccccccccccccccc}
\toprule
  & & \multicolumn{5}{c}{Gemma-3}
 & \multicolumn{4}{c}{Llama 3.1/3.2}
 & \multicolumn{7}{c}{Qwen-2.5} \\
\cmidrule(lr){3-7}\cmidrule(lr){8-11}\cmidrule(lr){12-18}
Task & Metric
 & 270M & 1B & 4B & 12B & 27B
 & 1B & 3B & 8B & 70B
 & 0.5B & 1.5B & 3B & 7B & 14B & 32B & 72B \\
\midrule\midrule
\multirow{4}{*}{TriviaQA}
 & Baseline & 6.7 & 23.9 & 54.1 & 74.4 & 83.0 & 40.2 & 63.3 & 77.0 & 90.9 & 16.8 & 36.9 & 44.9 & 58.1 & 72.5 & 74.8 & 83.2 \\
 & Cueing & 29.0 & 52.5 & 75.5 & 86.5 & 92.2 & 68.1 & 81.6 & 89.2 & 94.9 & 46.9 & 65.6 & 70.3 & 78.6 & 86.0 & 86.9 & 91.6 \\
 & $\Delta_{\mathrm{pp}}$ & +22.3 & +28.6 & +21.4 & +12.1 & +9.2 & +27.9 & +18.3 & +12.2 & +4.0 & +30.1 & +28.7 & +25.4 & +20.4 & +13.5 & +12.2 & +8.4 \\
 & $n$ & 6{,}560 & 6{,}560 & 6{,}560 & 6{,}560 & 6{,}560 & 7{,}075 & 7{,}075 & 7{,}075 & 7{,}075 & 7{,}157 & 7{,}157 & 7{,}157 & 7{,}157 & 7{,}157 & 7{,}157 & 7{,}157 \\
\midrule
\multirow{4}{*}{NQ-Open}
 & Baseline & 3.3 & 11.7 & 24.2 & 33.7 & 43.9 & 20.5 & 33.0 & 40.0 & 53.9 & 9.0 & 16.8 & 20.3 & 27.8 & 37.3 & 37.5 & 44.6 \\
 & Cueing & 13.9 & 25.4 & 41.5 & 52.5 & 59.3 & 43.2 & 56.1 & 59.0 & 70.0 & 23.6 & 35.2 & 39.1 & 47.3 & 51.8 & 54.0 & 60.4 \\
 & $\Delta_{\mathrm{pp}}$ & +10.6 & +13.7 & +17.3 & +18.8 & +15.4 & +22.7 & +23.1 & +19.0 & +16.2 & +14.6 & +18.4 & +18.8 & +19.5 & +14.5 & +16.5 & +15.9 \\
 & $n$ & 3{,}283 & 3{,}283 & 3{,}283 & 3{,}283 & 3{,}283 & 3{,}312 & 3{,}312 & 3{,}312 & 3{,}312 & 3{,}388 & 3{,}388 & 3{,}388 & 3{,}388 & 3{,}388 & 3{,}388 & 3{,}388 \\
\midrule
\multirow{4}{*}{WebQ}
 & Baseline & 9.9 & 22.2 & 35.8 & 44.2 & 48.6 & 30.4 & 39.8 & 45.5 & 52.6 & 18.9 & 29.5 & 31.9 & 42.3 & 45.8 & 46.9 & 50.3 \\
 & Cueing & 25.2 & 37.0 & 53.1 & 58.8 & 61.9 & 50.3 & 56.5 & 58.2 & 64.2 & 38.3 & 48.9 & 52.1 & 55.8 & 57.9 & 58.5 & 63.2 \\
 & $\Delta_{\mathrm{pp}}$ & +15.3 & +14.8 & +17.3 & +14.6 & +13.3 & +19.8 & +16.7 & +12.7 & +11.6 & +19.3 & +19.4 & +20.2 & +13.6 & +12.1 & +11.6 & +13.0 \\
 & $n$ & 1{,}750 & 1{,}750 & 1{,}750 & 1{,}750 & 1{,}750 & 1{,}866 & 1{,}866 & 1{,}866 & 1{,}866 & 1{,}866 & 1{,}866 & 1{,}866 & 1{,}866 & 1{,}866 & 1{,}866 & 1{,}866 \\
\midrule
\multirow{4}{*}{MKQA-EN}
 & Baseline & 3.7 & 10.6 & 22.7 & 35.6 & 43.3 & 18.8 & 30.8 & 37.9 & 52.4 & 8.2 & 16.2 & 20.0 & 29.3 & 37.8 & 39.8 & 46.8 \\
 & Cueing & 17.5 & 30.1 & 47.0 & 57.6 & 63.4 & 50.8 & 58.7 & 61.7 & 72.6 & 27.5 & 40.7 & 45.4 & 52.4 & 57.1 & 59.9 & 65.1 \\
 & $\Delta_{\mathrm{pp}}$ & +13.9 & +19.5 & +24.2 & +22.1 & +20.1 & +32.1 & +28.0 & +23.8 & +20.3 & +19.3 & +24.4 & +25.4 & +23.1 & +19.2 & +20.1 & +18.3 \\
 & $n$ & 6{,}168 & 6{,}168 & 6{,}168 & 6{,}168 & 6{,}168 & 6{,}355 & 6{,}355 & 6{,}355 & 6{,}355 & 6{,}357 & 6{,}357 & 6{,}357 & 6{,}357 & 6{,}357 & 6{,}357 & 6{,}357 \\
\midrule
\multirow{4}{*}{MKQA-ES}
 & Baseline & 2.7 & 9.6 & 19.6 & 31.9 & 39.5 & 8.8 & 17.0 & 27.5 & 47.7 & 5.4 & 10.2 & 15.0 & 24.4 & 31.9 & 34.8 & 43.0 \\
 & Cueing & 15.1 & 26.1 & 41.8 & 53.6 & 60.6 & 37.1 & 48.7 & 54.3 & 69.9 & 19.0 & 31.1 & 37.1 & 45.9 & 50.6 & 55.0 & 60.5 \\
 & $\Delta_{\mathrm{pp}}$ & +12.5 & +16.4 & +22.1 & +21.7 & +21.1 & +28.3 & +31.6 & +26.7 & +22.1 & +13.6 & +21.0 & +22.1 & +21.5 & +18.7 & +20.2 & +17.5 \\
 & $n$ & 6{,}366 & 6{,}366 & 6{,}366 & 6{,}366 & 6{,}366 & 6{,}499 & 6{,}499 & 6{,}499 & 6{,}499 & 6{,}501 & 6{,}501 & 6{,}501 & 6{,}501 & 6{,}501 & 6{,}501 & 6{,}501 \\
\midrule
\multirow{4}{*}{MKQA-AR}
 & Baseline & 1.5 & 6.7 & 12.1 & 20.9 & 30.7 & 2.5 & 5.1 & 9.3 & 28.4 & 2.3 & 4.2 & 7.5 & 12.6 & 16.0 & 19.4 & 28.9 \\
 & Cueing & 5.2 & 9.0 & 18.3 & 28.1 & 37.1 & 14.4 & 19.5 & 26.1 & 43.6 & 6.5 & 10.0 & 13.1 & 17.9 & 22.8 & 25.8 & 34.3 \\
 & $\Delta_{\mathrm{pp}}$ & +3.7 & +2.3 & +6.1 & +7.2 & +6.4 & +11.9 & +14.4 & +16.8 & +15.2 & +4.2 & +5.8 & +5.5 & +5.2 & +6.8 & +6.4 & +5.4 \\
 & $n$ & 6{,}595 & 6{,}595 & 6{,}595 & 6{,}595 & 6{,}595 & 6{,}598 & 6{,}598 & 6{,}598 & 6{,}598 & 6{,}551 & 6{,}551 & 6{,}551 & 6{,}551 & 6{,}551 & 6{,}551 & 6{,}551 \\
\midrule
\multirow{4}{*}{MKQA-RU}
 & Baseline & 2.3 & 8.1 & 15.9 & 28.5 & 38.3 & 4.3 & 11.0 & 18.6 & 43.5 & 3.1 & 7.2 & 12.0 & 18.9 & 23.7 & 28.0 & 39.5 \\
 & Cueing & 6.0 & 12.9 & 24.2 & 36.0 & 44.9 & 17.9 & 24.9 & 34.4 & 52.4 & 7.9 & 13.9 & 17.3 & 24.9 & 30.9 & 34.2 & 43.2 \\
 & $\Delta_{\mathrm{pp}}$ & +3.7 & +4.8 & +8.3 & +7.5 & +6.6 & +13.6 & +14.0 & +15.7 & +8.9 & +4.8 & +6.7 & +5.3 & +6.0 & +7.2 & +6.1 & +3.7 \\
 & $n$ & 6{,}573 & 6{,}573 & 6{,}573 & 6{,}573 & 6{,}573 & 6{,}588 & 6{,}588 & 6{,}588 & 6{,}588 & 6{,}587 & 6{,}587 & 6{,}587 & 6{,}587 & 6{,}587 & 6{,}587 & 6{,}587 \\
\midrule
\multirow{4}{*}{MKQA-KO}
 & Baseline & 1.2 & 5.6 & 11.7 & 22.0 & 32.3 & 3.0 & 8.5 & 12.9 & 32.2 & 1.8 & 4.3 & 7.8 & 14.2 & 17.7 & 22.6 & 31.8 \\
 & Cueing & 7.2 & 12.0 & 23.4 & 35.0 & 44.7 & 15.8 & 22.3 & 31.7 & 50.3 & 6.9 & 12.2 & 16.4 & 21.5 & 26.7 & 31.3 & 40.3 \\
 & $\Delta_{\mathrm{pp}}$ & +5.9 & +6.4 & +11.6 & +13.0 & +12.4 & +12.8 & +13.8 & +18.8 & +18.2 & +5.2 & +7.9 & +8.7 & +7.3 & +9.0 & +8.7 & +8.5 \\
 & $n$ & 6{,}572 & 6{,}572 & 6{,}572 & 6{,}572 & 6{,}572 & 6{,}587 & 6{,}587 & 6{,}587 & 6{,}587 & 6{,}590 & 6{,}590 & 6{,}590 & 6{,}590 & 6{,}590 & 6{,}590 & 6{,}590 \\
\midrule
\multirow{4}{*}{MKQA-JA}
 & Baseline & 2.2 & 6.9 & 13.3 & 25.6 & 35.2 & 3.9 & 11.2 & 17.6 & 38.5 & 2.7 & 6.1 & 8.7 & 15.0 & 20.1 & 24.0 & 34.7 \\
 & Cueing & 7.1 & 14.7 & 26.3 & 38.6 & 48.0 & 18.9 & 26.3 & 37.2 & 57.6 & 9.6 & 15.6 & 19.5 & 26.1 & 30.3 & 34.4 & 44.7 \\
 & $\Delta_{\mathrm{pp}}$ & +4.9 & +7.7 & +13.0 & +12.9 & +12.8 & +15.0 & +15.0 & +19.5 & +19.2 & +6.9 & +9.4 & +10.8 & +11.2 & +10.2 & +10.4 & +10.0 \\
 & $n$ & 6{,}453 & 6{,}453 & 6{,}453 & 6{,}453 & 6{,}453 & 6{,}580 & 6{,}580 & 6{,}580 & 6{,}580 & 6{,}517 & 6{,}517 & 6{,}517 & 6{,}517 & 6{,}517 & 6{,}517 & 6{,}517 \\
\midrule
\multirow{4}{*}{MKQA-ZH}
 & Baseline & 2.6 & 7.3 & 14.7 & 25.7 & 34.5 & 5.2 & 11.0 & 17.1 & 38.0 & 5.4 & 9.7 & 12.1 & 19.1 & 24.6 & 28.0 & 38.9 \\
 & Cueing & 7.4 & 13.5 & 24.7 & 35.4 & 44.4 & 20.0 & 25.9 & 35.3 & 50.7 & 16.1 & 22.9 & 27.6 & 32.8 & 36.9 & 42.2 & 49.9 \\
 & $\Delta_{\mathrm{pp}}$ & +4.8 & +6.2 & +9.9 & +9.7 & +9.9 & +14.9 & +14.9 & +18.2 & +12.7 & +10.7 & +13.1 & +15.5 & +13.7 & +12.4 & +14.2 & +11.0 \\
 & $n$ & 6{,}272 & 6{,}272 & 6{,}272 & 6{,}272 & 6{,}272 & 6{,}446 & 6{,}446 & 6{,}446 & 6{,}446 & 6{,}201 & 6{,}201 & 6{,}201 & 6{,}201 & 6{,}201 & 6{,}201 & 6{,}201 \\
\bottomrule
\end{tabular}%
}
\caption{%
    Cueing improves over Baseline in every cell ($160$ of $160$) of the
    headline subset, questions whose gold answer has at least two tokens
    under the model's tokenizer. Baseline and First-Token Cueing judge
    accuracy (\%), absolute improvement $\Delta_{\mathrm{pp}}$, and the
    number of questions $n$ for all $16$ models on the three English QA
    benchmarks and the seven MKQA languages.%
}
\label{tab:L2-cueing-results}
\end{table*}

Table~\ref{tab:fixbreak} splits the English QA gains on $L \geq 2$ into the
Baseline errors that the cue corrects and the Baseline-correct answers that it
breaks. Every model corrects more than three times as many errors as it breaks,
and an exact McNemar test on these discordant pairs gives $p < 10^{-100}$ for
every model. The share of errors corrected rises with scale among the smallest
models and stays between 40 and 50\% from there on, while the share of correct
answers broken falls with scale.

\begin{figure*}[t]
    \centering
    \includegraphics[width=\textwidth]{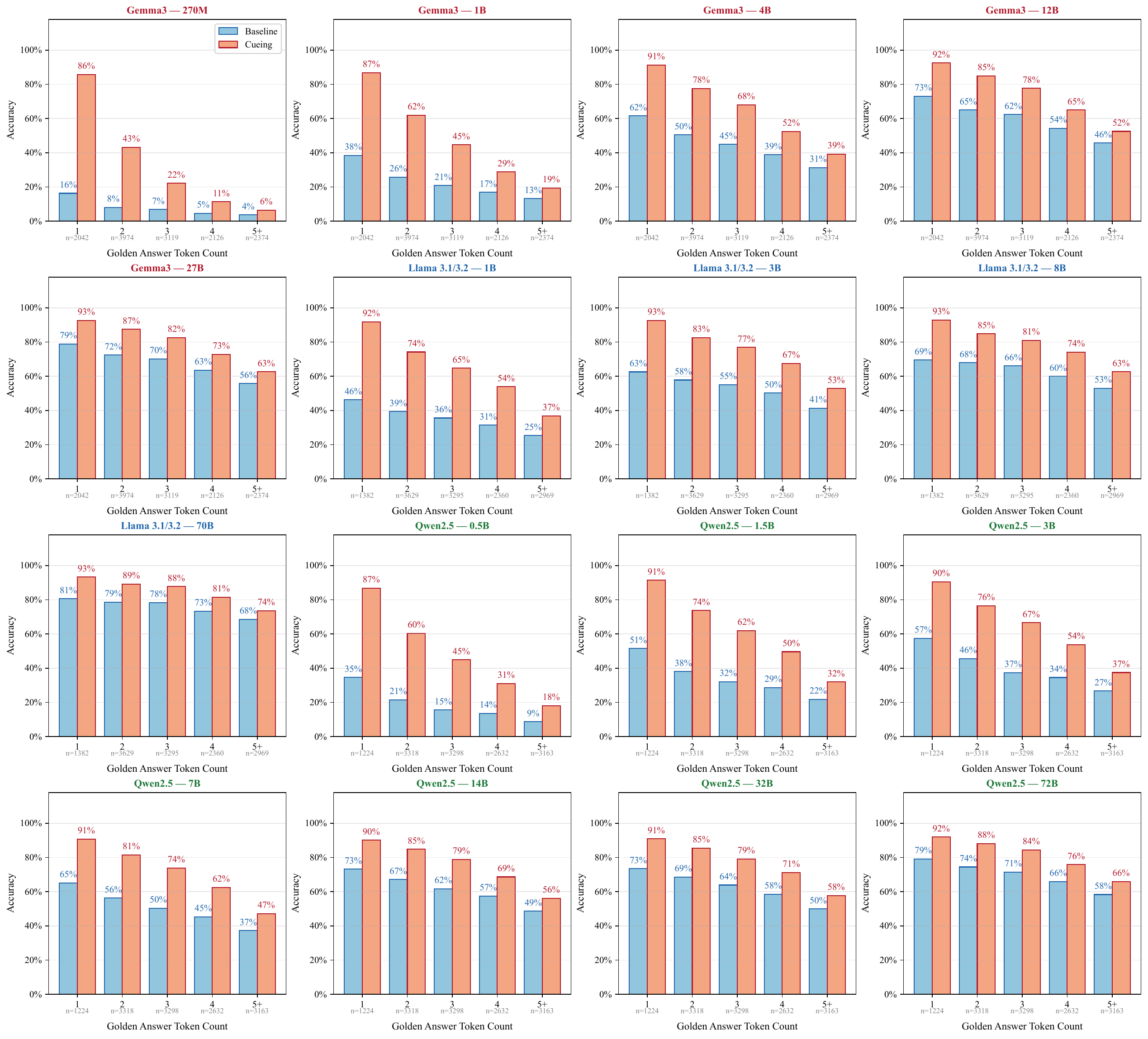}
    \caption{%
        Cueing exceeds Baseline at every gold answer length in every model.
        Baseline vs.\ First-Token Cueing accuracy by gold answer length
        (tokens) for all sixteen models, on all questions. Sample sizes
        appear below each pair.%
    }
    \label{fig:gold-length-bucket-all}
\end{figure*}

\subsection{Per-Model Accuracy by Gold Answer Length}
\label{app:gold-length-buckets}

Figure~\ref{fig:gold-length-bucket-all} breaks down Baseline and First-Token
Cueing accuracy by gold answer length for all sixteen models, on all questions
(Section~\ref{sec:length-analysis}). Two observations hold for every model.

First, Cueing improves over Baseline in every (model, gold length)
cell ($80$ of $80$), and the gain generally decreases with gold
answer length within each model. At gold length $1$, where the cue
alone is the gold answer, cueing improvements range from $+12$pp
to $+70$pp; at gold length $5{+}$, where the cue is the first of
five or more tokens, gains range from $+2$pp to $+12$pp. The
contraction with gold length is consistent with the cue providing
a fixed contribution of one correct token, whose relative weight
diminishes as the remaining answer grows longer.

Second, no model reaches $100\%$ cueing accuracy at gold length
$1$. The accuracy range at this position is $86\%$ (Gemma-3 270M)
to $93\%$ (four models tied: Gemma-3 27B, Llama-3.2 3B, Llama-3.1
8B, Llama-3.1 70B), and the gap to $100\%$ does not close even at
the largest scales we evaluate. When the cue is the whole answer, the
model must still stop after it, and it sometimes extends the cue into a
different answer instead (for Llama-3.1 8B, ``Colorado'' becomes ``Colorado
River'' and ``Cow'' becomes ``Cowpox'').

\begin{figure*}[t]
    \centering
    \includegraphics[width=\textwidth]{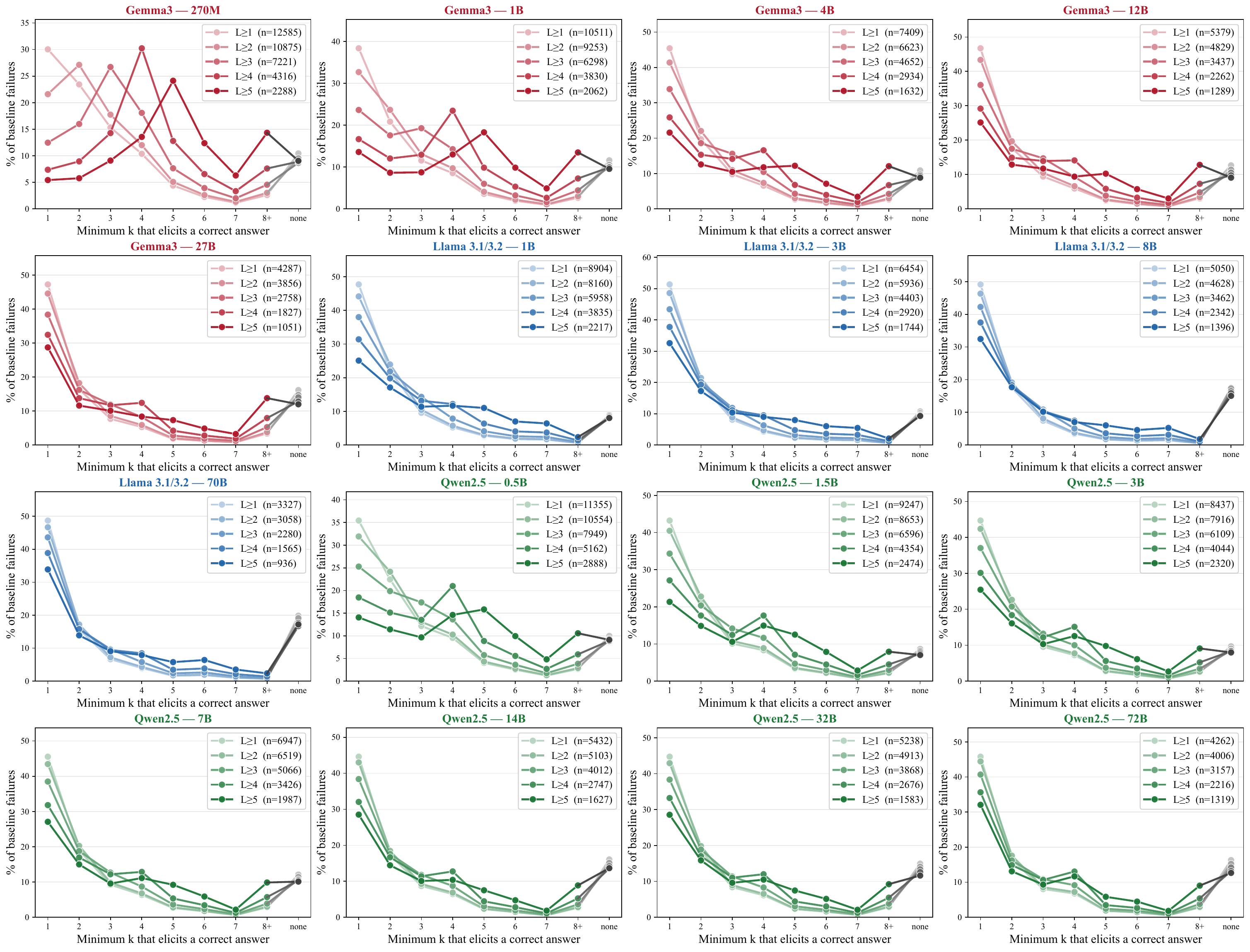}
    \caption{%
        The first token is the largest group at every answer length in 13 of
        the 16 models. Distribution of the shortest gold prefix $k$ that makes
        a Baseline error correct. Each curve $L \geq \ell$ restricts to gold
        answers of at least $\ell$ tokens; $8{+}$ collects $k \geq 8$, and none
        marks errors not recovered by any prefix up to the full gold answer.
        Sample sizes ($n$) appear in each panel's legend.%
    }
    \label{fig:k-distribution-all}
\end{figure*}

\subsection{Per-Model Minimum-$k$ Distribution}
\label{app:k-distribution}

Figure~\ref{fig:k-distribution-all} extends the analysis of
Section~\ref{sec:k-distribution} to all sixteen models, with the same
stratification by gold answer length ($L \geq \ell$ for
$\ell \in \{1, \ldots, 5\}$). Shares below are for $L \geq 2$.

First, $k = 1$ is the largest group in most models at every answer
length. The $k = 1$ share of Baseline errors ranges from $22\%$
(Gemma-3 270M) to $49\%$ (Llama-3.2 3B) and lies between $41\%$ and
$49\%$ for every model from 3B up; among errors recovered at any
prefix length, the share recovered at $k = 1$ ranges from $24\%$
(Gemma-3 270M) to $58\%$ (Llama-3.1 70B). In 13 of the 16 models,
$k = 1$ remains the largest group in every length stratum, including
gold answers of five or more tokens ($L \geq 5$). The exceptions are
the three models with the lowest Baseline accuracy: in Gemma-3 270M
from $L \geq 2$ on, and in Gemma-3 1B and Qwen-2.5 0.5B from
$L \geq 4$ on, the largest group is $k = \ell$, which for the
shortest answers in the stratum is the whole answer. These models
often recover a long answer only when given all of it.

Second, the share of errors not recovered by any prefix up to the
full gold answer (the none group) is larger in the largest model of
each family than in the smallest: from $10\%$ at Gemma-3 270M to
$15\%$ at Gemma-3 27B, from $8\%$ at Llama-3.2 1B to $19\%$ at
Llama-3.1 70B, and from $9\%$ at Qwen-2.5 0.5B to $15\%$ at
Qwen-2.5 72B. The share recovered only by longer prefixes shrinks
with scale, while the share recovered by the first token stays
roughly constant or rises (Table~\ref{tab:fixbreak}). Even in the
largest models the unrecovered residual remains a minority: $81$ to
$85\%$ of Baseline errors are recovered by some gold prefix. Because longer gold prefixes reveal
more of the answer, recovery at large $k$ should not be read as
evidence of recall; the analyses of Section~\ref{sec:recognition} concern the $k = 1$ cue.

\begin{figure*}[t]
    \centering
    \includegraphics[width=\textwidth]{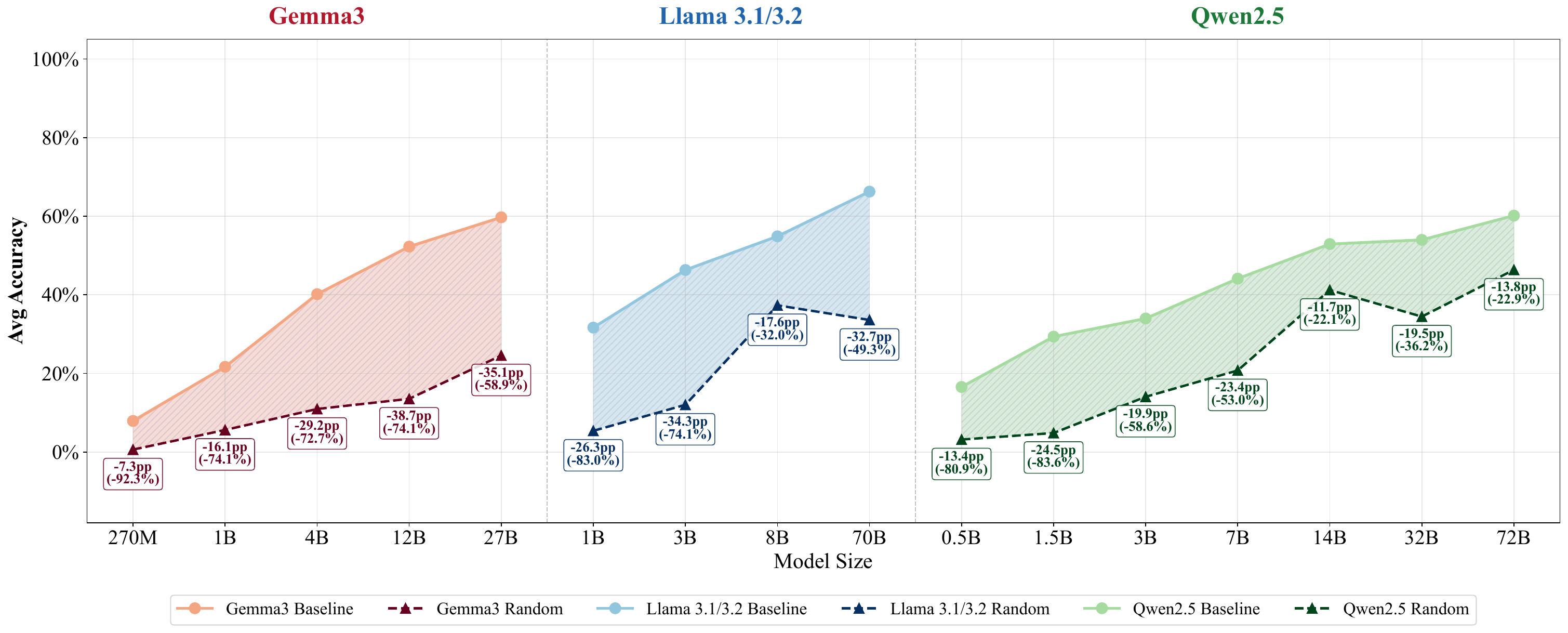}
    \caption{%
        A random first token lowers accuracy in every model. Average
        accuracy over TriviaQA, NQ-Open, and WebQuestions, Baseline (solid)
        vs.\ Random-Token (dashed), for all $16$ models. Annotations give the
        absolute (pp) and relative (\%) change.%
    }
    \label{fig:random-avg}
\end{figure*}

\begin{table*}[t]
\centering
\footnotesize
\setlength{\tabcolsep}{3pt}
\renewcommand{\arraystretch}{1.0}
\resizebox{\textwidth}{!}{%
\begin{tabular}{llccccccccccccccccc}
\toprule
  & & \multicolumn{5}{c}{Gemma-3}
 & \multicolumn{4}{c}{Llama 3.1/3.2}
 & \multicolumn{7}{c}{Qwen-2.5} \\
\cmidrule(lr){3-7}\cmidrule(lr){8-11}\cmidrule(lr){12-18}
Task & Metric
 & 270M & 1B & 4B & 12B & 27B
 & 1B & 3B & 8B & 70B
 & 0.5B & 1.5B & 3B & 7B & 14B & 32B & 72B \\
\midrule\midrule
\multirow{3}{*}{TriviaQA}
 & Baseline               & 8.5  & 26.9 & 56.6 & 75.7 & 83.9 & 41.5 & 64.3 & 77.2 & 90.9 & 18.8 & 39.0 & 46.9 & 59.5 & 73.5 & 75.6 & 83.8 \\
 & Random                 & 0.5  & 5.4  & 13.5 & 19.6 & 33.9 & 6.4  & 14.5 & 53.1 & 40.9 & 3.1  & 5.8  & 19.2 & 27.0 & 57.3 & 48.9 & 63.3 \\
 & $\Delta_{\mathrm{pp}}$ & $-8.0$ & $-21.5$ & $-43.1$ & $-56.1$ & $-50.0$ & $-35.1$ & $-49.8$ & $-24.1$ & $-50.0$ & $-15.7$ & $-33.2$ & $-27.7$ & $-32.4$ & $-16.2$ & $-26.7$ & $-20.4$ \\
\midrule
\multirow{3}{*}{NQ-Open}
 & Baseline               & 3.9  & 12.7 & 25.5 & 35.2 & 44.8 & 20.9 & 33.3 & 40.4 & 54.0 & 9.6  & 17.4 & 21.1 & 28.9 & 38.0 & 38.0 & 45.3 \\
 & Random                 & 0.3  & 3.8  & 7.2  & 9.7  & 18.4 & 4.0  & 9.6  & 27.5 & 29.2 & 2.1  & 3.0  & 8.4  & 14.1 & 29.3 & 24.2 & 35.5 \\
 & $\Delta_{\mathrm{pp}}$ & $-3.5$ & $-8.9$ & $-18.3$ & $-25.5$ & $-26.4$ & $-16.9$ & $-23.7$ & $-12.9$ & $-24.8$ & $-7.5$ & $-14.5$ & $-12.7$ & $-14.8$ & $-8.8$ & $-13.8$ & $-9.8$ \\
\midrule
\multirow{3}{*}{WebQ}
 & Baseline               & 11.3 & 25.4 & 38.3 & 45.8 & 50.4 & 32.6 & 41.3 & 47.0 & 53.8 & 21.3 & 31.6 & 33.9 & 44.0 & 47.2 & 48.3 & 51.4 \\
 & Random                 & 1.0  & 7.6  & 12.1 & 11.3 & 21.4 & 5.8  & 11.9 & 31.3 & 30.7 & 4.3  & 5.8  & 14.5 & 21.2 & 37.1 & 30.2 & 40.3 \\
 & $\Delta_{\mathrm{pp}}$ & $-10.3$ & $-17.8$ & $-26.2$ & $-34.5$ & $-29.0$ & $-26.8$ & $-29.4$ & $-15.6$ & $-23.2$ & $-17.0$ & $-25.9$ & $-19.4$ & $-22.8$ & $-10.1$ & $-18.1$ & $-11.1$ \\
\bottomrule
\end{tabular}%
}
\caption{%
    A random first token lowers accuracy in every (model, task) cell
    ($48$ of $48$), by $22.7$pp on average. Baseline and Random-Token
    accuracy (\%) and the absolute change $\Delta_{\mathrm{pp}}$ for all
    $16$ models on the three English QA benchmarks.%
}
\label{tab:random-per-task}
\end{table*}

\section{Random-Token Control}
\label{app:random-control}
This control tests whether the cueing gain requires the gold
first token or arises from prefilling any token. We replace
the gold token $t_{1}^{\star}$ in $\mathbf{c}^{\text{cue}}$
with a token sampled uniformly at random from the model's
vocabulary $\mathcal{V}$; the conditioning context, decoding
policy, and evaluation protocol are otherwise identical to
First-Token Cueing (Section~\ref{sec:first-token-cueing}).

Figure~\ref{fig:random-avg} reports Baseline and Random-Token
accuracy averaged across the three English open-domain QA
benchmarks. Per-task results appear in
Table~\ref{tab:random-per-task}.

\paragraph{Random tokens degrade accuracy in every model.}
All $16$ models lose accuracy under Random-Token relative to
Baseline, with average drops spanning $-7.3$pp on Gemma-3 270M
(where Baseline is already near floor) to $-38.7$pp on
Gemma-3 12B. The degradation holds in every (model, task) cell
($48$ of $48$). Whereas First-Token Cueing raises average QA accuracy by
about $18$pp (Table~\ref{tab:full-cueing-results}), a random
token at the same position lowers it by $22.7$pp on average.

Prefilling by itself therefore lowers accuracy, and the cueing gain requires
the cue to match the gold answer. This is consistent with encoding specificity
\citep{tulving1973encoding}: a cue helps only when it matches the target, and a
non-matching cue commits the model to an unrelated continuation.

\begin{table}[!t]
\centering
\resizebox{\columnwidth}{!}{%
\begin{tabular}{lcccccc}
\toprule
 & \multicolumn{3}{c}{Exact Match (\%)} & \multicolumn{3}{c}{Token F1} \\
\cmidrule(lr){2-4}\cmidrule(lr){5-7}
Model & Base & Cue & $\Delta$ & Base & Cue & $\Delta$ \\
\midrule\midrule
Gemma-3 270M  & 3.10 & 25.44 & +22.34 & 10.12 & 43.86 & +33.73 \\
Gemma-3 1B    & 10.47 & 30.72 & +20.25 & 18.40 & 48.07 & +29.67 \\
Gemma-3 4B    & 27.42 & 46.56 & +19.15 & 37.12 & 63.01 & +25.90 \\
Gemma-3 12B   & 39.71 & 55.94 & +16.23 & 49.52 & 71.51 & +21.99 \\
Gemma-3 27B   & 43.66 & 55.95 & +12.29 & 55.16 & 72.08 & +16.91 \\
\midrule
Llama-3.2 1B  & 21.13 & 46.68 & +25.55 & 30.39 & 61.96 & +31.57 \\
Llama-3.2 3B  & 32.57 & 52.24 & +19.67 & 42.70 & 67.51 & +24.81 \\
Llama-3.1 8B  & 37.82 & 51.65 & +13.84 & 49.01 & 67.20 & +18.19 \\
Llama-3.1 70B & 48.72 & 58.23 & +9.51  & 60.70 & 73.70 & +13.00 \\
\midrule
Qwen-2.5 0.5B & 7.21 & 29.74 & +22.52 & 15.79 & 45.18 & +29.39 \\
Qwen-2.5 1.5B & 18.88 & 42.88 & +24.00 & 27.43 & 57.67 & +30.24 \\
Qwen-2.5 3B   & 22.29 & 44.58 & +22.28 & 31.19 & 59.38 & +28.18 \\
Qwen-2.5 7B   & 27.50 & 46.75 & +19.25 & 37.84 & 61.98 & +24.14 \\
Qwen-2.5 14B  & 32.32 & 46.00 & +13.67 & 43.32 & 60.77 & +17.45 \\
Qwen-2.5 32B  & 37.42 & 50.30 & +12.88 & 48.44 & 65.61 & +17.17 \\
Qwen-2.5 72B  & 42.37 & 54.67 & +12.30 & 53.74 & 69.32 & +15.58 \\
\bottomrule
\end{tabular}%
}
\caption{Cueing improves exact match and token-level F1 for every model. Strict
alias-aware EM and token-level F1 (max over aliases; $0$--$100$ scale), Baseline
vs.\ First-Token Cueing, averaged over TriviaQA, NQ-Open, and WebQuestions (all
questions).}
\label{tab:lexical}
\end{table}

\section{Lexical Metrics: Exact Match and Token-Level F1}
\label{app:lexical}

We recompute the main results with two deterministic metrics applied directly
to the stored generations: strict
alias-aware Exact Match after SQuAD normalization (lowercasing, removing
punctuation and articles, collapsing whitespace) and SQuAD token-level F1,
taking the maximum over the gold answer and its aliases
\citep{rajpurkar2016squad}. Table~\ref{tab:lexical} reports both, averaged over
TriviaQA, NQ-Open, and WebQuestions (all questions). Cueing improves both metrics
for every model: EM by $+9.5$ to $+25.6$pp and F1 by $+13.0$ to $+33.7$ points.
Strict EM is lower than judge accuracy in absolute
terms because it rejects correct answers phrased with extra words.

On MKQA, we use the normalization of the official MKQA evaluation
\citep{longpre2021mkqa}, which also removes Spanish and Arabic articles and splits
Chinese and Japanese into characters. On questions whose gold answer has at least
two tokens, cueing improves both metrics in all $112$ (model, language) cells
(Table~\ref{tab:lexical-mkqa}). Because Chinese and Japanese are scored by
character, their F1 is not comparable with that of the other languages.

\begin{table}[!htb]
\centering
\resizebox{\columnwidth}{!}{%
\begin{tabular}{lcccccc}
\toprule
 & \multicolumn{3}{c}{Exact Match (\%)} & \multicolumn{3}{c}{Token F1} \\
\cmidrule(lr){2-4}\cmidrule(lr){5-7}
Language & Base & Cue & $\Delta$ & Base & Cue & $\Delta$ \\
\midrule\midrule
EN & 18.3 & 39.2 & $+20.9$ & 26.6 & 53.9 & $+27.4$ \\
ES & 13.3 & 33.5 & $+20.2$ & 19.2 & 46.9 & $+27.7$ \\
RU & 8.1 & 17.0 & $+8.9$ & 12.4 & 25.5 & $+13.1$ \\
AR & 4.3 & 13.2 & $+8.9$ & 7.8 & 23.0 & $+15.3$ \\
JA & 3.7 & 9.5 & $+5.8$ & 25.1 & 50.8 & $+25.7$ \\
ZH & 2.7 & 8.6 & $+5.8$ & 20.1 & 46.4 & $+26.3$ \\
KO & 3.5 & 11.6 & $+8.1$ & 5.4 & 19.8 & $+14.4$ \\
\bottomrule
\end{tabular}%
}
\caption{Cueing improves exact match and token-level F1 in every MKQA language.
Mean over the $16$ models on questions whose gold answer has at least two tokens,
with the official MKQA normalization (Chinese and Japanese F1 over characters).
Cueing improves both metrics in all $112$ (model, language) cells.}
\label{tab:lexical-mkqa}
\end{table}

\section{Additional Analyses of Cue Information}
\label{app:narrowing}

This appendix details the checks cited in Sections~\ref{sec:results-oqa} and~\ref{sec:recognition}. Unless stated otherwise, they use Llama-3.1 8B and 70B on the three English QA benchmarks (pooled). The oracle stratifications (Appendices~\ref{app:oracle-em}--\ref{app:ambiguity}) use strict EM; the cue controls and the thinking-based split use the LLM judge.

\subsection{Oracle under Exact Match}
\label{app:oracle-em}

\begin{table}[!t]
\centering
\resizebox{\columnwidth}{!}{%
\begin{tabular}{llrcccc}
\toprule
Model & Stratum & $n$ & Base & Cue & Oracle & Cue $-$ Oracle \\
\midrule\midrule
\multicolumn{7}{l}{Gold answers of at least two tokens} \\
\midrule
Llama-3.1 8B  & IN     & 7{,}620 & 62.2 & 75.6 & 77.3 & $-1.7$ \\
              & NOT-IN & 4{,}633 & 28.5 & 42.4 & 35.4 & $+7.0$ \\
Llama-3.1 70B & IN     & 8{,}149 & 73.5 & 81.4 & 82.5 & $-1.1$ \\
              & NOT-IN & 4{,}104 & 40.9 & 49.1 & 45.1 & $+4.0$ \\
\midrule
\multicolumn{7}{l}{All questions} \\
\midrule
Llama-3.1 8B  & IN     & 8{,}632 & 63.0 & 77.1 & 78.5 & $-1.3$ \\
              & NOT-IN & 5{,}003 & 27.8 & 43.9 & 34.8 & $+9.0$ \\
Llama-3.1 70B & IN     & 9{,}218 & 74.3 & 82.4 & 83.3 & $-0.9$ \\
              & NOT-IN & 4{,}417 & 39.4 & 50.0 & 43.9 & $+6.0$ \\
\bottomrule
\end{tabular}%
}
\caption{The oracle comparison of Table~\ref{tab:oracle} holds under strict EM.
Cueing vs.\ the oracle over the model's own first-token candidates (accuracy,
\%; three English benchmarks pooled), on questions with at least two gold tokens
and on all questions.}
\label{tab:oracle-em}
\end{table}

Table~\ref{tab:oracle-em} reports the comparison of Table~\ref{tab:oracle} over
questions under strict EM. The pattern matches the judge: cueing stays below the
oracle on IN questions and exceeds it on NOT-IN questions. Excluding single-token
answers, for which the cue is the whole answer, reduces the NOT-IN excess but
leaves it positive. The stratifications below use all questions.

\subsection{Candidate-Set Size}
\label{app:setsize}

\begin{table}[!t]
\centering
\resizebox{\columnwidth}{!}{%
\begin{tabular}{llrrccc}
\toprule
 & & & \multicolumn{2}{c}{Cue $-$ Base} & \multicolumn{2}{c}{Cue $-$ Oracle} \\
\cmidrule(lr){4-5}\cmidrule(lr){6-7}
Stratum & $|C|$ & $n$ (8B / 70B) & 8B & 70B & 8B & 70B \\
\midrule\midrule
IN     & 1   & 3{,}221 / 5{,}044 & $-0.2$ & $+0.0$ & $-0.3$ & $-0.0$ \\
       & 2   & 1{,}550 / 1{,}917 & $+8.7$ & $+9.4$ & $-1.7$ & $-1.0$ \\
       & 3--4 & 1{,}034 / 943 & $+18.9$ & $+19.8$ & $-3.6$ & $-3.4$ \\
       & 5--8 & 721 / 492 & $+22.1$ & $+22.8$ & $-2.1$ & $-3.5$ \\
       & 9+  & 2{,}106 / 822 & $+34.9$ & $+32.1$ & $-1.4$ & $-2.0$ \\
\midrule
NOT-IN & 1   & 1{,}509 / 1{,}911 & $+5.2$ & $+2.0$ & $+5.3$ & $+1.9$ \\
       & 2   & 825 / 852 & $+16.7$ & $+14.0$ & $+10.1$ & $+7.0$ \\
       & 3--4 & 730 / 631 & $+23.4$ & $+16.8$ & $+12.7$ & $+6.5$ \\
       & 5--8 & 597 / 390 & $+20.4$ & $+14.6$ & $+10.2$ & $+8.0$ \\
       & 9+  & 1{,}342 / 633 & $+21.8$ & $+23.1$ & $+10.1$ & $+15.3$ \\
\bottomrule
\end{tabular}%
}
\caption{Stratification by candidate-set size $|C|$ (pp; pooled over
the three English QA benchmarks). The gross gain grows with $|C|$, but the excess
over the oracle does not: it is
$\leq 0$ in $10/10$ IN cells and $> 0$ in $10/10$ NOT-IN cells, including
$|C| = 1$, where there is nothing to select among.}
\label{tab:setsize}
\end{table}

The gross gain grows with the number of candidates the cue rules out
(Table~\ref{tab:setsize}): on IN questions it rises from about $0$pp at
$|C| = 1$ to $+32$--$35$pp at $|C| \geq 9$. The excess over the oracle does not
depend on set size: it is
$\leq 0$ in all IN cells and $> 0$ in all NOT-IN cells. In particular, for
NOT-IN questions with a single candidate, the model's candidate set holds one
non-gold token and there is nothing to select among, yet cueing still exceeds
the oracle ($+5.3$ and $+1.9$pp).

\subsection{Prefix Ambiguity}
\label{app:ambiguity}

\begin{table}[!t]
\centering
\resizebox{\columnwidth}{!}{%
\begin{tabular}{llrcc}
\toprule
 & & & \multicolumn{2}{c}{Cue $-$ Oracle} \\
\cmidrule(lr){4-5}
Stratum & Ambiguity & $n$ (8B / 70B) & 8B & 70B \\
\midrule\midrule
IN     & 1 (unique) & 880 / 946 & $-0.3$ & $+0.0$ \\
       & 2--5  & 1{,}834 / 1{,}972 & $-0.2$ & $-0.6$ \\
       & 6--20 & 1{,}045 / 1{,}129 & $-1.1$ & $-0.9$ \\
       & 21+   & 567 / 587 & $-2.5$ & $-2.7$ \\
\midrule
NOT-IN & 1 (unique) & 748 / 682 & $+8.4$ & $+8.1$ \\
       & 2--5  & 936 / 798 & $+15.8$ & $+14.3$ \\
       & 6--20 & 444 / 360 & $+15.3$ & $+10.3$ \\
       & 21+   & 334 / 314 & $+0.6$ & $+1.6$ \\
\bottomrule
\end{tabular}%
}
\caption{Stratification by prefix ambiguity: the number of distinct
normalized gold answers in the dataset that share the gold first token (token
length $\geq 3$; questions whose first token is the full answer excluded; pp,
pooled over the three English QA benchmarks). The excess over the oracle peaks
at intermediate ambiguity, not for unique prefixes.}
\label{tab:ambiguity}
\end{table}

If the cue helped in proportion to how much it reveals about the answer, first
tokens unique to one answer should benefit most. We measure the ambiguity of the
gold first token as the number of distinct normalized gold
answers in the dataset that share it (restricting to first tokens of at least
three characters and excluding questions whose first token is the full answer).
Table~\ref{tab:ambiguity} shows a different ordering on NOT-IN questions: unique prefixes exceed the oracle by $+8$pp,
whereas prefixes shared by $2$--$20$ answers, which cannot identify the answer
on their own, exceed it by $+10$ to $+16$pp. The excess vanishes only for
cues shared by $21$ or more answers, which carry almost no information. Completing the
correct answer from a prefix that does not identify it requires the model to
use the question, which is more consistent with recall than with prefix
informativeness. We note that this ambiguity is measured within each dataset,
so it is a proxy for how many plausible answers a prefix admits in general.

\subsection{Cue Controls}
\label{app:controls}

\begin{table}[!t]
\centering
\resizebox{\columnwidth}{!}{%
\begin{tabular}{llrccc}
\toprule
Model & Dataset & $n$ & Base & Gold cue & Plausible wrong cue \\
\midrule\midrule
Llama-3.1 8B  & TriviaQA & 5{,}262 & 66.3 & 86.1 & 51.7 \\
              & NQ-Open  & 3{,}280 & 35.4 & 58.2 & 27.4 \\
              & WebQ     & 1{,}725 & 38.9 & 53.9 & 30.6 \\
\midrule
Llama-3.1 70B & TriviaQA & 3{,}991 & 82.7 & 91.8 & 70.4 \\
              & NQ-Open  & 2{,}927 & 44.6 & 65.8 & 37.4 \\
              & WebQ     & 1{,}569 & 42.5 & 57.1 & 35.6 \\
\bottomrule
\end{tabular}%
}
\caption{Plausible-but-incorrect first-token cue (judge accuracy, \%):
the highest-probability non-gold token of the model's own top-$p$ set,
continued greedily; $n$ covers questions with at least one non-gold candidate.
It restricts the output space through the same mechanism as the gold cue, yet
reduces accuracy below Baseline. Strict EM shows the same ordering 
(Table~\ref{tab:plausible-wrong-em}).}
\label{tab:plausible-wrong}
\end{table}

\begin{table}[!t]
\centering
\resizebox{\columnwidth}{!}{%
\begin{tabular}{lrcccc}
\toprule
Dataset & $n$ & Base & Output prefilling & Prefill $-$ Base & Cue $-$ Base \\
\midrule\midrule
TriviaQA & 7{,}993 & 77.2 & 78.0 & $+0.8$ & $+13.0$ \\
NQ-Open  & 3{,}610 & 40.6 & 41.3 & $+0.7$ & $+20.8$ \\
WebQ     & 2{,}032 & 47.5 & 46.8 & $-0.7$ & $+12.7$ \\
\bottomrule
\end{tabular}%
}
\caption{Generic output prefilling \citep{cappelletti2025improving} vs.\
gold First-Token Cueing on open-ended QA (Llama-3.1 8B, judge accuracy, \%,
all questions). Output prefilling prepends ``Given the question, my answer is: ''
(no gold information). The two interventions were run as separate passes, so
their Baselines differ marginally (cueing-run Baselines: $77.2$ / $40.4$ /
$47.0$); each delta is computed against its own run.}
\label{tab:prefill}
\end{table}

\paragraph{Plausible but incorrect first tokens.}
A token that restricts the output space but does not match the gold answer
isolates the effect of restriction. We cue the model with the
highest-probability non-gold token in its own top-$p$ set, which in practice is
a same-type competitor of the gold answer. Although this token narrows the
output space through the same mechanism as the gold token, it reduces judge
accuracy by $7$--$15$pp below Baseline (Table~\ref{tab:plausible-wrong}),
whereas the gold token raises it by $9$--$23$pp on the same questions.
Restriction to a plausible start therefore does not by itself help; what helps
is the start matching the answer. Together with the random-token control
(Appendix~\ref{app:random-control}), this shows that the benefit depends on the
content of the cue rather than on generic prefilling effects.

\paragraph{Generic output prefilling.}
\citet{cappelletti2025improving} improve first-token predictions in
multiple-choice QA by prefilling a generic prefix. Applying their approach to
open-ended QA (Table~\ref{tab:prefill}), the prefix ``Given the question, my
answer is: '' yields no meaningful gain ($+0.8$ / $+0.7$ / $-0.7$pp), whereas
the gold first token gains $+13.0$ / $+20.8$ / $+12.7$pp under the same judge.
A prefix that only fixes the format of the answer does not account for the
cueing gain.

\paragraph{Exact match.}
Table~\ref{tab:plausible-wrong-em} complements Table~\ref{tab:plausible-wrong}
with strict EM. The judge is the primary metric for this control because EM is
structurally handicapped when a completion starts with a wrong token (the model
sometimes recovers the correct entity later in the response, which EM rejects);
both metrics show the same ordering.

\begin{table}[!h]
\centering
\resizebox{\columnwidth}{!}{%
\begin{tabular}{llrccc}
\toprule
Model & Dataset & $n$ & Base EM & Gold-cue EM & Plausible-wrong-cue EM \\
\midrule\midrule
Llama-3.1 8B  & TriviaQA & 5{,}262 & 56.0 & 78.9 & 38.5 \\
              & NQ-Open  & 3{,}280 & 18.9 & 38.4 & 10.3 \\
              & WebQ     & 1{,}725 & 13.0 & 23.1 & 5.0 \\
\midrule
Llama-3.1 70B & TriviaQA & 3{,}991 & 72.8 & 85.6 & 57.4 \\
              & NQ-Open  & 2{,}927 & 25.4 & 44.5 & 14.7 \\
              & WebQ     & 1{,}569 & 13.5 & 22.1 & 6.8 \\
\bottomrule
\end{tabular}%
}
\caption{Plausible-but-incorrect first-token cue, strict EM (\%).}
\label{tab:plausible-wrong-em}
\end{table}

\subsection{Relation to Hidden Knowledge: Thinking-Recoverable Failures}
\label{app:thinking-overlap}

\begin{table}[!t]
\centering
\resizebox{\columnwidth}{!}{%
\begin{tabular}{llrcc}
\toprule
Model & Thinking pass@16 & $n$ & Share of failures & Cue recovery \\
\midrule\midrule
Llama-3.1 8B  & recoverable     & 2{,}745 & 59.3\% & 59.9\% \\
              & not recoverable & 1{,}883 & 40.7\% & 26.6\% \\
\midrule
Llama-3.1 70B & recoverable     & 1{,}442 & 47.2\% & 61.0\% \\
              & not recoverable & 1{,}616 & 52.8\% & 33.8\% \\
\bottomrule
\end{tabular}%
}
\caption{Cueing recovery vs.\ thinking-based recovery. Baseline
failures (judge; three English QA benchmarks pooled; gold answers of at least
two tokens) split by whether
inference-time thinking (pass@16 over $16$ CoT samples) recovers them, the
criterion \citet{calderon2026empty} use to mark facts as encoded but not
directly recallable. Cueing recovery concentrates on the failures this
criterion marks as encoded.}
\label{tab:thinking-overlap}
\end{table}

Probing studies show that correct answers can often be recovered from a model's
internal states \citep{burns2023discovering, azaria2023internal,
orgad2025llms}. We connect our results to this literature through
\citet{calderon2026empty}, who show that inference-time thinking resolves recall
failures for facts that are encoded in the model but not directly recalled. We
use their criterion, thinking pass@16, as an independent behavioral marker that
the knowledge behind a Baseline failure is encoded, and compare cueing
recovery between failures with and without this marker. On gold answers of at least two tokens, cueing recovers about $60\%$ of
thinking-recoverable failures but only $27$--$34\%$ of the rest, $1.8$--$2.3\times$
more (Table~\ref{tab:thinking-overlap}); on all questions the ratio is
$1.7$--$2.1\times$. Equivalently, $77\%$ (8B) and $62\%$ (70B) of the
cue-recoverable failures are thinking-recoverable, against $59\%$ and $47\%$ of
all failures. The two
interventions, one supplying a gold token and one using only the model's own
sampled reasoning, thus largely converge on the same failures.

\begin{table*}[!t]
\centering
\footnotesize
\setlength{\tabcolsep}{3pt}
\resizebox{\textwidth}{!}{%
\begin{tabular}{lcccccccccccccccc}
\toprule
  & \multicolumn{5}{c}{Gemma-3} & \multicolumn{4}{c}{Llama 3.1/3.2} & \multicolumn{7}{c}{Qwen-2.5} \\
\cmidrule(lr){2-6}\cmidrule(lr){7-10}\cmidrule(lr){11-17}
First-token length & 270M & 1B & 4B & 12B & 27B & 1B & 3B & 8B & 70B & 0.5B & 1.5B & 3B & 7B & 14B & 32B & 72B \\
\midrule\midrule
1 char   & +9.5 & +12.8 & +14.5 & +10.7 & +8.2 & +21.2 & +17.0 & +12.4 & +6.0 & +16.9 & +17.3 & +18.2 & +16.7 & +10.8 & +9.8 & +8.1 \\
2 chars  & +11.5 & +15.4 & +16.2 & +13.0 & +10.0 & +24.2 & +17.1 & +13.6 & +9.5 & +20.5 & +20.1 & +17.7 & +15.3 & +10.1 & +10.2 & +10.1 \\
3 chars  & +17.8 & +17.0 & +18.5 & +15.8 & +12.3 & +25.7 & +19.6 & +13.0 & +9.5 & +22.4 & +24.9 & +23.3 & +19.7 & +14.5 & +12.6 & +13.0 \\
4+ chars & +31.5 & +25.5 & +22.3 & +19.5 & +14.7 & +29.9 & +23.2 & +16.2 & +12.1 & +29.2 & +31.2 & +27.2 & +23.4 & +17.2 & +17.0 & +16.6 \\
\bottomrule
\end{tabular}%
}
\caption{Cueing gain by the character length of the gold first token
(strict EM, Cue $-$ Baseline in pp, averaged over TriviaQA, NQ-Open, and
WebQuestions). All $64$ cells are positive, and the 4+-character bucket is the
largest for every model.}
\label{tab:gain-by-toklen}
\end{table*}

\section{Tokenizer and Script Effects on Cue Strength}
\label{app:tokenizer}

A ``first token'' carries a different amount of information depending on the
tokenizer and the script: it may be a whole word, a subword, a single character,
or an informative entity fragment. We quantify how much of the answer the first
token carries under each tokenizer and script.

\paragraph{Across tokenizers.}
On the English benchmarks, the mean gold first token has $3.84$, $3.16$, and
$3.06$ characters under the Gemma-3, Llama-3, and Qwen-2.5 tokenizers, and
$20$--$31\%$ of questions receive a single-character first token
(Table~\ref{tab:tokenizer-qa}). Stratifying the cueing gain by first-token
length (Table~\ref{tab:gain-by-toklen}), all $64$ (model, length) cells are
positive, so the effect is present even when the cue is a single character. The
gain grows with the length of the cue in the same way under all three
tokenizers, and the 4+-character bucket is the largest for every model. A more
informative cue thus yields a larger gain, as expected if part of the gain comes
from the information the cue carries; Section~\ref{sec:recognition} separates this
contribution from recall.

\paragraph{Across languages.}
Tokenizer effects are much larger across the seven MKQA languages
(Figure~\ref{fig:tokenizer} and Table~\ref{tab:tokenizer}). In the non-Latin scripts (RU, AR, JA, ZH, KO),
$38$--$94\%$ of first tokens are a single character, and answers span more
tokens ($4.9$--$6.5$ vs.\ $3.6$--$4.7$ for EN and ES), so the $k = 1$ cue covers
a smaller fraction of the answer. This is consistent with the smaller gains in
non-Latin languages (Section~\ref{sec:results-mkqa}), and with Llama, whose
tokenizer yields the longest first tokens in those scripts, showing the largest
multilingual gains.

\begin{table}[!t]
\centering
\resizebox{\columnwidth}{!}{%
\begin{tabular}{lccccccc}
\toprule
 & mean & \multicolumn{4}{c}{first-token length (chars)} & tok & tokens \\
\cmidrule(lr){3-6}
Family & chars & 1 & 2 & 3 & 4+ & = ans. & / answer \\
\midrule\midrule
Gemma-3  & 3.84 & 19.9\% & 10.1\% & 22.9\% & 47.1\% & 15.0\% & 3.21 \\
Llama-3  & 3.16 & 25.6\% & 16.2\% & 26.2\% & 32.1\% & 10.1\% & 3.41 \\
Qwen-2.5 & 3.06 & 31.3\% & 14.6\% & 22.0\% & 32.1\% & 9.0\%  & 3.60 \\
\bottomrule
\end{tabular}%
}
\caption{Gold first-token statistics per tokenizer family, pooled over
TriviaQA, NQ-Open, and WebQuestions ($n = 13{,}635$). ``tok = ans.'' is the
share of questions whose first token is already the complete answer.}
\label{tab:tokenizer-qa}
\end{table}

\begin{table}[!t]
\centering
\resizebox{\columnwidth}{!}{%
\begin{tabular}{lccc}
\toprule
  & First-token chars & 1-char first token (\%) & Tokens per answer \\
Data & (G / L / Q) & (G / L / Q) & (G / L / Q) \\
\midrule\midrule
English QA & 3.84 / 3.16 / 3.06 & 19.9 / 25.6 / 31.3 & 3.21 / 3.41 / 3.60 \\
\midrule
MKQA EN & 3.23 / 3.14 / 2.64 & 38.3 / 19.5 / 46.8 & 4.04 / 3.60 / 4.34 \\
ES & 3.01 / 2.88 / 2.39 & 39.5 / 21.9 / 49.3 & 4.23 / 3.96 / 4.69 \\
RU & 1.57 / 1.78 / 1.25 & 64.9 / 56.3 / 87.7 & 5.68 / 5.45 / 6.48 \\
AR & 1.63 / 1.99 / 1.70 & 58.5 / 38.1 / 54.3 & 5.49 / 5.08 / 5.82 \\
JA & 1.58 / 1.79 / 1.39 & 60.3 / 49.7 / 71.3 & 5.61 / 5.61 / 6.45 \\
ZH & 1.23 / 1.62 / 1.31 & 84.1 / 64.1 / 77.5 & 5.46 / 5.18 / 5.36 \\
KO & 1.17 / 1.63 / 1.13 & 91.0 / 65.2 / 93.5 & 5.37 / 4.89 / 5.88 \\
\bottomrule
\end{tabular}%
}
\caption{Gold first-token statistics by tokenizer family (G: Gemma-3, L: Llama-3,
Q: Qwen-2.5): mean length of the first token in characters, share of
single-character first tokens, and mean gold answer length in tokens. English QA
pools TriviaQA, NQ-Open, and WebQuestions; the other rows are MKQA languages. Plotted in
Figure~\ref{fig:tokenizer}.}
\label{tab:tokenizer}
\end{table}

\begin{table*}[!t]
\centering
\footnotesize
\resizebox{\textwidth}{!}{%
\begin{tabular}{lcccccccccccccccc}
\toprule
  & \multicolumn{4}{c}{Llama-3.1 8B ($4{,}608$ eligible)} & \multicolumn{4}{c}{Llama-3.1 70B ($3{,}995$)} & \multicolumn{4}{c}{Qwen-2.5 7B ($5{,}481$)} & \multicolumn{4}{c}{Gemma-3 12B ($6{,}324$)} \\
\cmidrule(lr){2-5}\cmidrule(lr){6-9}\cmidrule(lr){10-13}\cmidrule(lr){14-17}
Caliper & pairs & cov. & pref. & adv. & pairs & cov. & pref. & adv. & pairs & cov. & pref. & adv. & pairs & cov. & pref. & adv. \\
\midrule\midrule
1 nat  & 873 & 37.9 & 65.6 & +0.37 & 469 & 23.5 & 63.4 & +0.34 & 1{,}109 & 40.5 & 63.4 & +0.31 & 513 & 16.2 & 60.1 & +0.08 \\
2 nats & 1{,}171 & 50.8 & 65.5 & +0.37 & 710 & 35.5 & 64.5 & +0.32 & 1{,}488 & 54.3 & 64.0 & +0.31 & 843 & 26.7 & 59.4 & +0.07 \\
3 nats & 1{,}371 & 59.5 & 66.7 & +0.38 & 878 & 44.0 & 65.2 & +0.31 & 1{,}681 & 61.3 & 64.4 & +0.31 & 1{,}074 & 34.0 & 60.5 & +0.07 \\
5 nats & 1{,}615 & 70.1 & 68.4 & +0.38 & 1{,}174 & 58.8 & 68.3 & +0.32 & 1{,}951 & 71.2 & 65.6 & +0.30 & 1{,}372 & 43.4 & 60.8 & +0.07 \\
none   & 2{,}297 & 99.7 & 74.8 & +0.41 & 1{,}989 & 99.6 & 76.5 & +0.36 & 2{,}707 & 98.8 & 68.4 & +0.31 & 3{,}147 & 99.5 & 65.8 & +0.07 \\
\bottomrule
\end{tabular}%
}
\caption{Crossed-cue control under relaxed prior matching. cov.: \% of
eligible questions in a pair; pref.: \% of questions whose correct-cue
continuation is more confident (no prior stratification); adv.: confidence
advantage of the correct start at equal prior (regression intercept, nats of
entropy). The advantage is stable as coverage grows.}
\label{tab:crossed-sweep}
\end{table*}

\section{Crossed-Cue Control: Details}
\label{app:crossed}

\paragraph{Answer types.}
We label every question's expected answer type with \texttt{gpt-oss-120b} (low
reasoning effort, greedy), given the question and the gold answer, using a fixed
two-level taxonomy. The coarse types, used for pairing, are \textsc{person}
($4{,}107$ questions), \textsc{thing} ($3{,}440$), \textsc{location} ($2{,}692$),
\textsc{work} ($971$), \textsc{date} ($937$), \textsc{organization} ($927$), and
\textsc{number} ($497$); $64$ questions labeled \textsc{other} are excluded. Each
coarse type has fine subtypes (e.g., city, country, natural feature for
\textsc{location}), which we use only as a robustness check. All but one of the
$13{,}635$ labels were parsed. The labels were not verified by humans, and
\textsc{thing} is a broad category.

\paragraph{Eligibility and pairing.}
For each model, a question is eligible if its gold first token is NOT-IN the
model's own top-$p$ ($p = 0.95$) first-step candidate set, its gold answer has
at least two tokens under the model's tokenizer, and its type is not
\textsc{other}. We pair eligible questions within the same dataset and coarse
type. A pair $(i, j)$ is admissible if the two gold first tokens differ, the two
gold answers differ, and neither token can begin any gold answer or alias of the
other question (checked by string prefix). Using exact first-step log-probabilities
of all gold first tokens under each question, we choose pairs by greedy
minimum-cost matching with cost
$|\log p(t_j \mid q_i) - \log p(t_i \mid q_i)| + |\log p(t_i \mid q_j) - \log p(t_j \mid q_j)|$,
so that each question's incorrect start is as close as possible in prior
probability to its correct start. Each question is used at most once. Optionally,
a caliper restricts both prior gaps to at most $c$ nats.

\paragraph{Confidence and statistics.}
For each of the four (question, cue) cells of a pair we generate a greedy
continuation (up to $50$ tokens) and record per-token entropies, approximated
from the top-$20$ log-probabilities plus a uniform tail over the remaining
vocabulary. Confidence is the negative mean entropy over the continuation. A
question counts as preferring the correct start if its correct-cue continuation
has lower mean entropy than its incorrect-cue continuation. We report Wilson
intervals and exact two-sided binomial tests against $50\%$, and the intercept of
a regression of the per-question confidence difference on the prior gap
(the confidence advantage of the correct start at equal prior). As a
manipulation check, \texttt{gpt-oss-120b} judges whether each output is an
answer of the right kind for the question, regardless of correctness.

\paragraph{Coverage.}
Table~\ref{tab:crossed-sweep} relaxes the caliper from $1$ nat to none. Coverage
rises to $99\%$ of eligible questions. The raw preference rises as well, because
looser matching admits pairs in which the correct start was more probable, but
the confidence advantage at equal prior is stable within each model. The main
text therefore reports the matched-prior stratum of the uncalipered pairs
(Table~\ref{tab:crossed}). The smaller absolute advantage for Gemma-3 12B
reflects its lower entropy scale: its first-step distributions are often
saturated.

\paragraph{Robustness (Llama-3.1 8B).}
With a $1$-nat caliper, the share of pairs whose summed confidence difference
favors the correct cues (chance $50\%$) is $72.2\%$ with coarse types ($873$
pairs), $68.8\%$ with fine types ($618$), $70.9\%$ when both incorrect-cue outputs
are judged to be of the right type ($625$), $71.4\%$ when both cue tokens begin
with a letter or digit ($791$), and $69.7\%$ when the Baseline answered both
questions incorrectly ($544$). Across coarse types it ranges from $64.2\%$
(\textsc{date}) to $80.9\%$ (\textsc{work}). Four confidence measures (mean
entropy, mean log-probability, next-token entropy, and mean entropy of the first
three tokens) give the same conclusion.

\section{Gold-Free Search over First Characters}
\label{app:e2}

\paragraph{Procedure.}
For Llama-3.1 8B on the three English QA benchmarks ($13{,}635$ questions), we
compute the exact first-step distribution and, for each of the $36$ characters
A--Z and 0--9, take the most probable vocabulary token whose text starts with that
character (case-insensitive, ignoring leading spaces). We prefill this token,
continue greedily, and obtain $36$ candidate answers per question. A selector
picks one candidate using only the model's own scores: the total log-probability
of the answer (first token plus continuation) or the log-probability of the
continuation alone. We compare against greedy decoding, beam search (width $8$),
and self-consistency ($16$ samples at temperature $0.7$, top-$p$ $0.95$,
majority vote), and report oracle references that use the gold answer. All
numbers are strict EM pooled over the three benchmarks.

\paragraph{Constrained first token.}
Prefilling the gold answer's first character as a single-character token
breaks the tokenization the model expects: with this naive prefill, Llama-3.1 8B
loses $13.0$, $7.1$, and $7.8$pp of judge accuracy on TriviaQA, NQ-Open, and
WebQuestions, often switching to an entity whose canonical first token is a
single letter (e.g., George Carlin $\to$ Gore Vidal). Constraining the
first token to tokens that begin with the character avoids this and raises EM by
$+10.5$pp instead (Table~\ref{tab:e2}).

\begin{table}[!t]
\centering
\resizebox{\columnwidth}{!}{%
\begin{tabular}{lcc}
\toprule
 & All & Greedy first char.\ wrong \\
Method & ($n = 13{,}635$) & ($n = 6{,}055$) \\
\midrule\midrule
Greedy (Baseline)                     & 50.0 & 12.1 \\
Beam search (width 8)                 & 53.0 & 16.9 \\
Self-consistency (16 samples)         & 51.6 & 15.0 \\
36 characters, total log-prob         & 52.2 & 17.2 \\
36 characters, continuation log-prob  & 45.7 & 20.5 \\
\midrule
\multicolumn{3}{l}{Oracle references (use the gold answer)} \\
Gold first character (constrained)    & 60.5 & 35.6 \\
Any of the 36 candidates              & 64.2 & 41.7 \\
Gold first token (First-Token Cueing) & 64.9 & 44.7 \\
\bottomrule
\end{tabular}%
}
\caption{Gold-free search over first characters (Llama-3.1 8B, strict
EM, \%). Searching over all first characters with the model's own scores gains
about as much as beam search; the correct candidate is usually in the set but
not selected.}
\label{tab:e2}
\end{table}

\paragraph{Results.}
Selecting by total log-probability gains $+2.1$pp, slightly less than beam search
($+3.0$pp), although the candidate set contains a correct answer for $64.2\%$ of
questions. Selecting by continuation confidence alone fixes more errors when the
greedy first character is wrong ($+8.4$pp) but abandons correct answers elsewhere
($-4.4$pp overall). On questions whose greedy first character is wrong, the
correct-character continuation has lower mean entropy than the median of the
other $35$ continuations in $82.7\%$ of questions, but lower than the
continuation of the model's own (wrong) character in only $39.9\%$. The model can
thus distinguish a correct start from arbitrary alternatives, but not from the
start it already prefers. Between two starts of equal prior probability, its
confidence favors the correct one (Section~\ref{sec:crossed}); among many starts
that include its own preferred one, the preferred start usually wins. Greedy
chain-of-thought does not find the correct start either
(Figure~\ref{fig:thinking-comparison}).

\section{First-Token Entropy and Gold-Token Rank Across Models}
\label{app:rank}

For each of the $16$ models, we compute the first-step distribution of every
question in the three English QA benchmarks with vLLM, using the top-$1{,}000$
log-probabilities plus a uniform tail for the remaining mass; for Llama-3.1 8B
this approximation agrees with the exact entropy at Pearson $r = 0.999$. The
recomputed greedy first token matches the stored Baseline output in $96$--$99\%$
of questions ($83\%$ for Llama-3.2 1B, whose chat template inserts the current
date).

Table~\ref{tab:auroc} reports the AUROC with which first-token entropy predicts
whether the Baseline answer is correct. Table~\ref{tab:rank} groups Baseline
errors by the rank of the gold first token: recovery falls from $65.5\%$ when the
gold token was ranked 2nd--5th to $30.3\%$ beyond rank $1{,}000$.

\begin{table}[!htb]
\centering
\resizebox{\columnwidth}{!}{%
\begin{tabular}{lclc}
\toprule
Model & AUROC & Model & AUROC \\
\midrule\midrule
Gemma-3 270M & 0.633 & Llama-3.1 70B & 0.832 \\
Gemma-3 1B & 0.668 & Qwen-2.5 0.5B & 0.729 \\
Gemma-3 4B & 0.757 & Qwen-2.5 1.5B & 0.801 \\
Gemma-3 12B & 0.778 & Qwen-2.5 3B & 0.758 \\
Gemma-3 27B & 0.763 & Qwen-2.5 7B & 0.739 \\
Llama-3.2 1B & 0.773 & Qwen-2.5 14B & 0.704 \\
Llama-3.2 3B & 0.819 & Qwen-2.5 32B & 0.761 \\
Llama-3.1 8B & 0.809 & Qwen-2.5 72B & 0.752 \\
\bottomrule
\end{tabular}%
}
\caption{First-token entropy predicts Baseline correctness (AUROC of
negative first-token entropy for judge correctness; three English QA benchmarks
pooled).}
\label{tab:auroc}
\end{table}

\begin{table}[!htb]
\centering
\resizebox{\columnwidth}{!}{%
\begin{tabular}{lcc}
\toprule
Rank of gold first token & Share of errors (\%) & Fixed by the cue (\%) \\
\midrule\midrule
1 (the greedy first token) & 7.4 & 5.5 \\
2--5 & 21.6 & 65.5 \\
6--50 & 30.0 & 50.7 \\
51--1{,}000 & 27.4 & 37.5 \\
$>1{,}000$ & 13.6 & 30.3 \\
\bottomrule
\end{tabular}%
}
\caption{Recovery by the rank of the gold first token under the Baseline
first-step distribution (Baseline errors, three English QA benchmarks; mean over
$16$ models). Errors whose gold first token was the model's runner-up are fixed
most often; when the greedy first token already is the gold token, the cue leaves
the input unchanged.}
\label{tab:rank}
\end{table}

\section{Results on All Questions}
\label{app:all-questions}

\begin{table*}[tp]
\centering
\footnotesize
\setlength{\tabcolsep}{3pt}
\renewcommand{\arraystretch}{1.0}
\resizebox{\textwidth}{!}{%
\begin{tabular}{llccccccccccccccccc}
\toprule
  & & \multicolumn{5}{c}{Gemma-3}
 & \multicolumn{4}{c}{Llama 3.1/3.2}
 & \multicolumn{7}{c}{Qwen-2.5} \\
\cmidrule(lr){3-7}\cmidrule(lr){8-11}\cmidrule(lr){12-18}
Task & Metric
 & 270M & 1B & 4B & 12B & 27B
 & 1B & 3B & 8B & 70B
 & 0.5B & 1.5B & 3B & 7B & 14B & 32B & 72B \\
\midrule\midrule
\multirow{4}{*}{TriviaQA}
 & Baseline               & 8.5  & 26.9 & 56.6 & 75.7 & 83.9 & 41.5 & 64.3 & 77.2 & 90.9 & 18.8 & 39.0 & 46.9 & 59.5 & 73.5 & 75.6 & 83.8 \\
 & Cueing                 & 40.3 & 59.9 & 79.5 & 88.7 & 93.4 & 71.4 & 83.5 & 90.2 & 95.4 & 51.7 & 68.9 & 73.1 & 80.5 & 87.2 & 88.1 & 92.3 \\
 & $\Delta_{\mathrm{pp}}$ & +31.8 & +33.0 & +22.9 & +12.9 & +9.4 & +29.9 & +19.3 & +13.0 & +4.6 & +33.0 & +29.9 & +26.2 & +21.1 & +13.7 & +12.5 & +8.6 \\
 & $\Delta\%$             & +373.4 & +122.5 & +40.4 & +17.0 & +11.3 & +72.2 & +29.9 & +16.9 & +5.0 & +175.4 & +76.8 & +55.9 & +35.5 & +18.7 & +16.5 & +10.2 \\
\midrule
\multirow{4}{*}{NQ-Open}
 & Baseline               & 3.9  & 12.7 & 25.5 & 35.2 & 44.8 & 20.9 & 33.3 & 40.4 & 54.0 & 9.6  & 17.4 & 21.1 & 28.9 & 38.0 & 38.0 & 45.3 \\
 & Cueing                 & 19.4 & 29.6 & 45.0 & 55.2 & 61.4 & 46.8 & 58.5 & 61.2 & 71.3 & 26.8 & 38.1 & 41.4 & 49.4 & 53.3 & 55.4 & 61.8 \\
 & $\Delta_{\mathrm{pp}}$ & +15.6 & +17.0 & +19.5 & +20.0 & +16.6 & +25.9 & +25.2 & +20.8 & +17.3 & +17.2 & +20.7 & +20.3 & +20.5 & +15.3 & +17.4 & +16.5 \\
 & $\Delta\%$             & +405.0 & +133.6 & +76.5 & +56.8 & +37.2 & +124.1 & +75.7 & +51.4 & +31.9 & +179.5 & +118.9 & +96.5 & +71.0 & +40.3 & +45.7 & +36.5 \\
\midrule
\multirow{4}{*}{WebQ}
 & Baseline               & 11.3 & 25.4 & 38.3 & 45.8 & 50.4 & 32.6 & 41.3 & 47.0 & 53.8 & 21.3 & 31.6 & 33.9 & 44.0 & 47.2 & 48.3 & 51.4 \\
 & Cueing                 & 30.8 & 41.4 & 55.6 & 60.8 & 63.6 & 52.0 & 58.0 & 59.7 & 65.1 & 40.7 & 51.0 & 53.9 & 57.1 & 59.0 & 59.7 & 63.9 \\
 & $\Delta_{\mathrm{pp}}$ & +19.4 & +16.0 & +17.3 & +15.0 & +13.2 & +19.4 & +16.6 & +12.7 & +11.2 & +19.4 & +19.3 & +20.0 & +13.1 & +11.7 & +11.4 & +12.5 \\
 & $\Delta\%$             & +171.7 & +63.2 & +45.2 & +32.8 & +26.2 & +59.7 & +40.2 & +27.0 & +20.8 & +91.4 & +61.1 & +59.1 & +29.8 & +24.8 & +23.6 & +24.4 \\
\midrule
\multirow{4}{*}{MKQA-EN}
 & Baseline               & 5.2  & 12.8 & 25.0 & 37.5 & 45.0 & 20.1 & 31.8 & 38.9 & 53.3 & 9.6  & 17.8 & 21.4 & 30.8 & 39.1 & 41.1 & 48.0 \\
 & Cueing                 & 24.1 & 35.2 & 51.0 & 60.8 & 66.0 & 53.4 & 60.8 & 63.4 & 74.0 & 31.3 & 43.8 & 48.3 & 54.8 & 59.1 & 61.9 & 66.8 \\
 & $\Delta_{\mathrm{pp}}$ & +18.8 & +22.4 & +26.0 & +23.3 & +21.0 & +33.3 & +29.0 & +24.5 & +20.7 & +21.6 & +26.0 & +26.9 & +24.0 & +20.1 & +20.8 & +18.8 \\
 & $\Delta\%$             & +360.6 & +174.4 & +104.0 & +62.2 & +46.6 & +165.6 & +91.0 & +62.9 & +38.8 & +224.9 & +145.8 & +125.7 & +78.1 & +51.4 & +50.7 & +39.2 \\
\midrule
\multirow{4}{*}{MKQA-ES}
 & Baseline               & 3.4  & 10.7 & 20.8 & 33.1 & 40.5 & 9.5  & 17.7 & 28.2 & 48.4 & 6.2  & 11.1 & 15.7 & 25.2 & 32.6 & 35.6 & 43.9 \\
 & Cueing                 & 19.5 & 29.6 & 44.6 & 55.9 & 62.6 & 39.3 & 50.2 & 55.7 & 70.9 & 21.7 & 33.4 & 38.9 & 47.6 & 52.0 & 56.5 & 61.8 \\
 & $\Delta_{\mathrm{pp}}$ & +16.2 & +18.9 & +23.8 & +22.8 & +22.1 & +29.8 & +32.5 & +27.5 & +22.5 & +15.4 & +22.2 & +23.2 & +22.4 & +19.4 & +20.9 & +18.0 \\
 & $\Delta\%$             & +481.5 & +177.5 & +114.6 & +68.8 & +54.4 & +314.8 & +184.0 & +97.3 & +46.4 & +247.7 & +199.6 & +147.6 & +88.7 & +59.5 & +58.8 & +41.0 \\
\midrule
\multirow{4}{*}{MKQA-AR}
 & Baseline               & 2.2  & 7.4  & 12.9 & 21.6 & 31.2 & 3.4  & 5.9  & 10.2 & 29.2 & 3.2  & 4.9  & 8.3  & 13.5 & 16.9 & 20.3 & 29.9 \\
 & Cueing                 & 7.4  & 10.9 & 20.0 & 29.7 & 38.5 & 16.3 & 21.1 & 27.6 & 44.8 & 9.0  & 12.5 & 15.4 & 20.1 & 24.7 & 27.7 & 36.0 \\
 & $\Delta_{\mathrm{pp}}$ & +5.2 & +3.5 & +7.1 & +8.1 & +7.2 & +12.9 & +15.2 & +17.4 & +15.7 & +5.8 & +7.5 & +7.1 & +6.5 & +7.9 & +7.4 & +6.1 \\
 & $\Delta\%$             & +238.1 & +46.5 & +54.9 & +37.7 & +23.2 & +384.1 & +255.6 & +171.6 & +53.8 & +184.6 & +152.9 & +86.3 & +48.1 & +46.7 & +36.4 & +20.4 \\
\midrule
\multirow{4}{*}{MKQA-RU}
 & Baseline               & 3.1  & 8.8  & 16.7 & 29.2 & 38.9 & 5.1  & 11.7 & 19.4 & 44.1 & 3.9  & 7.9  & 12.7 & 19.5 & 24.5 & 28.8 & 40.3 \\
 & Cueing                 & 8.4  & 15.0 & 26.1 & 37.7 & 46.4 & 19.9 & 26.6 & 35.9 & 53.5 & 10.0 & 16.0 & 19.3 & 26.6 & 32.4 & 35.8 & 44.6 \\
 & $\Delta_{\mathrm{pp}}$ & +5.3 & +6.2 & +9.4 & +8.4 & +7.5 & +14.8 & +14.9 & +16.5 & +9.5 & +6.1 & +8.1 & +6.6 & +7.1 & +7.9 & +7.0 & +4.3 \\
 & $\Delta\%$             & +172.7 & +70.8 & +56.4 & +28.9 & +19.2 & +291.0 & +127.6 & +85.2 & +21.5 & +156.8 & +102.3 & +52.2 & +36.3 & +32.2 & +24.4 & +10.6 \\
\midrule
\multirow{4}{*}{MKQA-KO}
 & Baseline               & 1.8  & 6.1  & 12.4 & 22.6 & 32.9 & 3.4  & 8.8  & 13.5 & 32.6 & 2.5  & 5.0  & 8.2  & 14.9 & 18.5 & 23.4 & 32.5 \\
 & Cueing                 & 9.4  & 14.1 & 25.3 & 36.5 & 46.0 & 17.7 & 24.0 & 33.1 & 51.4 & 8.9  & 14.2 & 18.2 & 23.2 & 28.2 & 32.7 & 41.6 \\
 & $\Delta_{\mathrm{pp}}$ & +7.6 & +7.9 & +12.8 & +14.0 & +13.1 & +14.4 & +15.2 & +19.6 & +18.9 & +6.4 & +9.1 & +10.0 & +8.3 & +9.7 & +9.4 & +9.1 \\
 & $\Delta\%$             & +414.5 & +129.4 & +103.0 & +61.9 & +39.9 & +423.6 & +173.2 & +145.1 & +57.9 & +256.8 & +180.9 & +121.0 & +55.8 & +52.8 & +40.2 & +27.9 \\
\midrule
\multirow{4}{*}{MKQA-JA}
 & Baseline               & 3.2  & 8.1  & 14.6 & 26.7 & 36.1 & 4.8  & 12.0 & 18.5 & 39.2 & 3.8  & 7.1  & 9.6  & 16.1 & 21.1 & 24.9 & 35.7 \\
 & Cueing                 & 10.7 & 17.7 & 29.2 & 41.0 & 49.9 & 20.9 & 28.1 & 38.7 & 58.7 & 12.3 & 18.2 & 22.0 & 28.5 & 32.3 & 36.4 & 46.4 \\
 & $\Delta_{\mathrm{pp}}$ & +7.5 & +9.6 & +14.6 & +14.3 & +13.8 & +16.1 & +16.1 & +20.2 & +19.5 & +8.5 & +11.1 & +12.4 & +12.4 & +11.1 & +11.5 & +10.7 \\
 & $\Delta\%$             & +233.3 & +118.4 & +99.9 & +53.5 & +38.4 & +331.8 & +134.7 & +108.8 & +49.9 & +220.8 & +155.9 & +129.8 & +76.7 & +52.7 & +46.2 & +30.0 \\
\midrule
\multirow{4}{*}{MKQA-ZH}
 & Baseline               & 4.1  & 9.1  & 16.8 & 27.4 & 35.8 & 6.5  & 12.0 & 18.4 & 38.7 & 7.5  & 11.9 & 14.1 & 21.2 & 26.6 & 30.1 & 40.7 \\
 & Cueing                 & 13.6 & 18.8 & 29.4 & 39.5 & 47.8 & 23.4 & 29.0 & 38.0 & 52.7 & 22.1 & 28.5 & 32.8 & 37.4 & 41.0 & 46.2 & 53.2 \\
 & $\Delta_{\mathrm{pp}}$ & +9.5 & +9.7 & +12.6 & +12.1 & +12.0 & +16.9 & +17.0 & +19.6 & +14.0 & +14.6 & +16.6 & +18.7 & +16.2 & +14.4 & +16.2 & +12.6 \\
 & $\Delta\%$             & +233.8 & +105.7 & +75.3 & +44.4 & +33.5 & +261.9 & +142.0 & +106.8 & +36.2 & +195.2 & +139.0 & +133.0 & +76.4 & +54.2 & +53.7 & +30.9 \\
\bottomrule
\end{tabular}%
}
\caption{%
    Cueing also improves over Baseline in every cell ($160$ of $160$) on
    all questions, including single-token gold answers. Baseline and
    First-Token Cueing judge accuracy (\%), absolute improvement
    $\Delta_{\mathrm{pp}}$, and relative improvement $\Delta\%$ for all $16$
    models on the three English QA benchmarks and the seven MKQA languages.%
}
\label{tab:full-cueing-results}
\end{table*}

\begin{figure*}[tp]
    \centering
    \includegraphics[width=\textwidth]{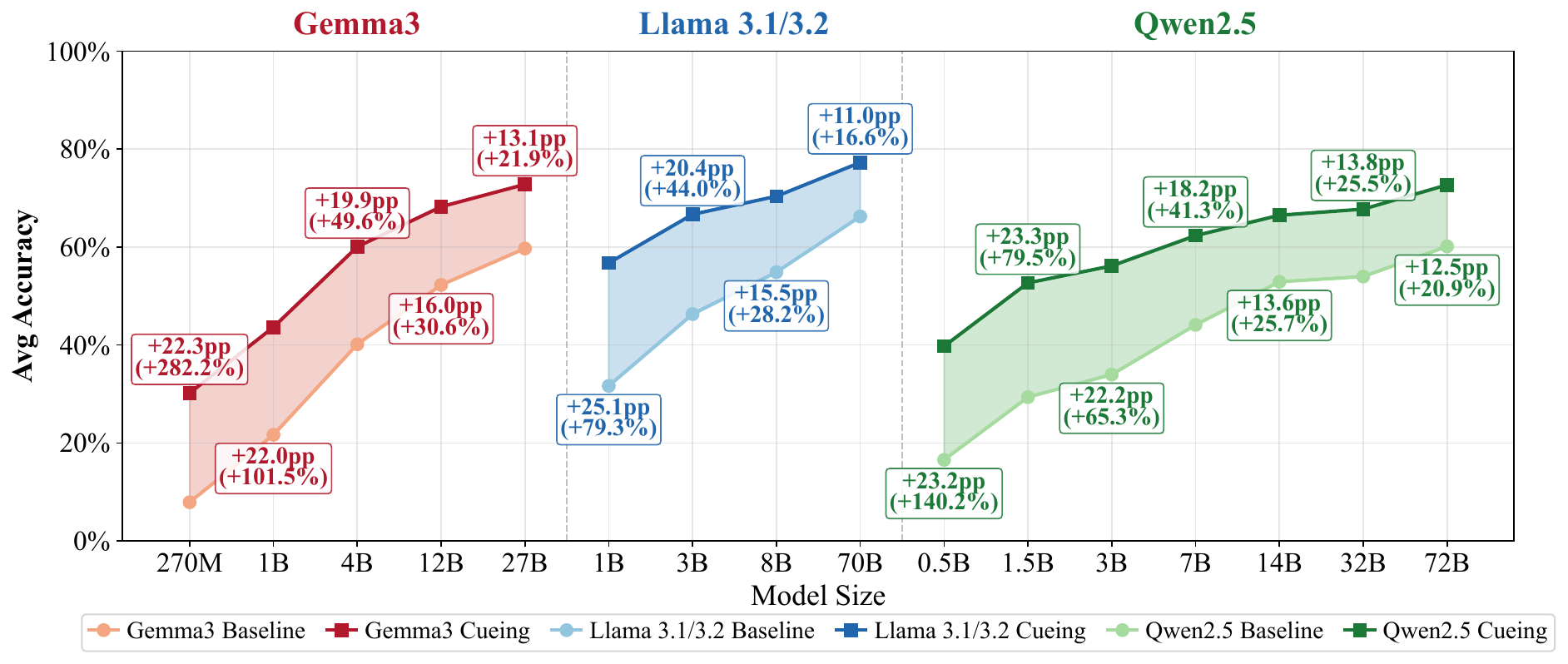}
    \caption{%
        All questions (including single-token answers).
        Counterpart of Figure~\ref{fig:cueing-avg}: average
        open-domain QA accuracy (TriviaQA, NQ-Open, WebQuestions),
        Baseline (light) vs.\ First-Token Cueing (dark).
        Annotations give absolute (pp) and relative (\%) gains.%
    }
    \label{fig:cueing-avg-all}
\end{figure*}

\begin{figure*}[tp]
    \centering
    \includegraphics[width=\textwidth]{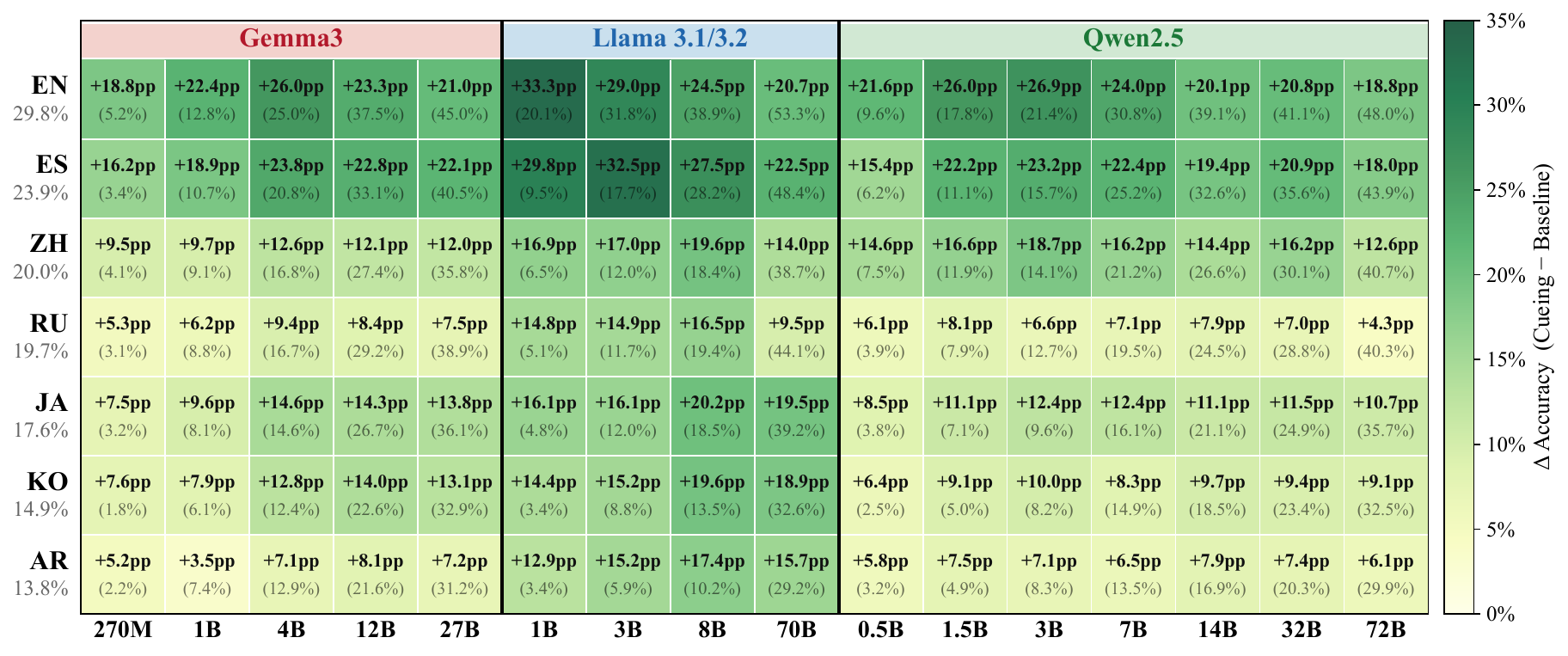}
    \caption{%
        All questions (including single-token answers).
        Counterpart of Figure~\ref{fig:mkqa-heatmap}: Cueing $-$
        Baseline accuracy gap (pp) on MKQA for $16$ models
        $\times$ $7$ languages; per-cell Baseline accuracy in
        parentheses. All $112$ cells are positive.%
    }
    \label{fig:mkqa-heatmap-all}
\end{figure*}

Table~\ref{tab:full-cueing-results} reports all questions, including single-token
gold answers, for which the cue is the complete answer. Cueing improves over
Baseline in all $160$ (model, task) cells. Excluding single-token answers reduces
the gains slightly, most for TriviaQA, which has the largest share of them
($13\%$ on average across models). Figures~\ref{fig:cueing-avg-all}
and~\ref{fig:mkqa-heatmap-all} show the main-text figures computed on all
questions.

\end{document}